\documentclass{article}

\usepackage[accsupp]{axessibility}  

    \usepackage[preprint]{neurips_2022}

\usepackage[utf8]{inputenc} 
\usepackage[T1]{fontenc}    
\usepackage{hyperref}       
\usepackage{url}            
\usepackage{booktabs}       
\usepackage{amsfonts}       
\usepackage{nicefrac}       
\usepackage{microtype}      
\usepackage{xcolor}         

\usepackage[pdftex]{graphicx}

\usepackage{amsmath}
\def\triangleq{\overset{\Delta}{=}}

\usepackage[capitalize]{cleveref}
\crefname{section}{Sec.}{Secs.}
\Crefname{section}{Section}{Sections}
\Crefname{table}{Table}{Tables}
\crefname{table}{Tab.}{Tabs.}

\newtheorem{claim}{Claim}
\crefname{claim}{Claim}{Claims}

\Crefname{algorithm}{Algorithm}{Algorithms}
\crefname{algorithm}{Algo.}{Algos.}

\usepackage{todonotes}

\newcommand{\specialcell}[2][c]{\begin{tabular}[#1]{@{}c@{}}#2\end{tabular}}
\newcommand{\specialcellleft}[2][l]{\begin{tabular}[#1]{@{}l@{}}#2\end{tabular}}

\newtheorem{proposition}{Proposition}
\newtheorem{definition}{Definition}

\usepackage{algorithm}
\usepackage{algpseudocode}

\usepackage{caption}
\usepackage{subcaption}

\title{Alias-Free Convnets: \\ Fractional Shift Invariance via Polynomial Activations}

\author{
      Hagay Michaeli \ \ \ \ \  
      Tomer Michaeli  \ \ \ \ \  
      Daniel Soudry \\
      Department of Electrical and Computer Engineering \\
      Technion, Haifa, Israel\\
    \texttt{\{hagaymichaeli, daniel.soudry\}@gmail.com, tomer.m@ee.technion.ac.il} 
}

\begin{document}

\maketitle

\begin{abstract}
Although CNNs are believed to be invariant to translations, recent works have shown this is not the case, due to aliasing effects that stem from downsampling layers. 
The existing architectural solutions to prevent aliasing are partial since they do not solve these effects, that originate in non-linearities.
We propose an extended anti-aliasing method that tackles both downsampling and non-linear layers, thus creating truly alias-free, shift-invariant CNNs\footnote{Our code is available at \href{https://github.com/hmichaeli/alias_free_convnets}{https://github.com/hmichaeli/alias\_free\_convnets/}.}. 
We show that the presented model is invariant to integer as well as fractional (i.e., sub-pixel) translations, thus outperforming other shift-invariant methods in terms of robustness to adversarial translations.
\end{abstract}

\section{Introduction}
Convolutional Neural Networks (CNNs) are the most common model in the image classification field. They were originally intended to have two properties:
\begin{enumerate}
    \item Shift-invariant output: when we spatially translate the input image, their output does not change. 
    \item Shift-equivariant representation: when we spatially translate the input image, their internal representation translates in the same way.
\end{enumerate}
Both these properties are thought to be beneficial for generalization (i.e., they are useful inductive biases), as we expect the image class not to change by an image translation, and its features to shift together with the image. Moreover, without the first property, the CNN might become vulnerable to adversarial attacks using image translations. Such attacks are real threats since they are very simple to execute in a ``black-box'' setting (where we do not know anything about the CNN). For example, consider a person trying to fool a CNN-based face scanner, by simply moving continuously until a face match is achieved.

It was commonly assumed that these useful properties were maintained since CNNs use only shift-equivariant operations: the convolution operation and component-wise non-linearities. However, CNN models typically also include downsampling operations such as pooling and strided convolution. Unfortunately, these operations violate equivariance, and this also leads to CNNs not being shift-invariant.
Specifically, \citet{Azulay2019WhyTransformations} have shown that shifting an input image by even one pixel can cause the output probability of a trained classifier to change significantly. This vulnerability can be further exploited in adversarial attacks, lowering classifiers’ accuracy by more than 20\% \citep{Engstrom2017ExploringRobustness}.
Later, \citet{Zhang2019MakingAgain} has shown that this problematic behavior stems from an aliasing effect, taking place in downsampling operations such as pooling and strided convolutions, and non-linear operations on the downsampled signals.

Previous works have shown an improvement in CNN invariance to translations using partial solutions that reduced aliasing. For example, \citet{Zhang2019MakingAgain} has suggested adding a low-pass filter before the downsampling operations. This approach has been shown to reduce aliasing caused by downsampling, thus improving shift-invariance, as well as accuracy and noise robustness.
\citet{Karras2021Alias-FreeNetworks} have addressed aliasing in the generator within generative adversarial networks (GANs). 
They have shown that without proper treatment, aliasing in GANs leads to a decoupling of the high-frequency features (texture) from the low-frequency content (structure) in the generated images, thus limiting their applicability in smooth video generation. 
To alleviate this issue, \citet{Karras2021Alias-FreeNetworks} extended the low-pass filter approach and suggested a solution for the implicit aliasing caused by non-linearities. 
 Their method wraps the component-wise non-linear operations by upsampling and downsampling layers in an attempt to mimic the effect of applying the non-linear operations in the continuous domain, where they theoretically do not cause aliasing. 

Yet, none of the previous solutions completely eliminates aliasing, thus their suggested CNN architectures are not guaranteed to be shift-invariant.
A different approach to shift-invariant CNNs was suggested by \citet{Chaman2020TrulyNetworks}. 
They have proposed to use downsampling operations that dynamically choose the subsampling grid using a shift-equivariant decision rule.
Although it does not solve the aliasing problem, nor guarantees shift-equivariant representations, this approach enables the creation of CNNs whose outputs are completely invariant to integer circular shifts. 
However, this approach does not lead to invariance to subpixel shifts, which are common in real-world applications. 
For example, consider a case where the CNN receives input from a camera with some finite resolution. 
If we continuously shift the camera with respect to the scene, then the resulting shift in the CNN's discretely sampled input would rarely be integer-valued. 

Considering the anti-aliasing approach again, the problem with the solution suggested by \citet{Karras2021Alias-FreeNetworks} is that the aliasing resulting from non-linearities that increase the signal's bandwidth indefinitely (such as ReLU) can be avoided only when they are used in a continuous domain (i.e., with infinite resolution), which is impractical. 
However, this problem can be solved by replacing such non-linearities with alternatives whose effect does not lead to an indefinite increase in the signal's bandwidth --- such as polynomials.

\paragraph{Polynomial activations}
Despite their ease of computation, polynomials are not considered promising candidates for activation functions. 
The main practical reason for this is that polynomial activations have large (super-linear) magnitudes compared to standard activations (e.g., ReLU) and thus typically cause training instability (e.g., exploding gradients) \citep{Gottemukkula2019POLYNOMIALFUNCTIONS}. 
There seems also to be a theoretical disadvantage since shallow feedforward neural networks with polynomial activation functions are not universal approximators \citep{Hornik1989MultilayerApproximators}.
However, this last issue may not be a serious disadvantage: \citet{Kidger2020UniversalNetworks} have shown that feedforward neural networks with polynomial activations can become universal approximators with sufficient depth --- a regime more relevant for modern CNNs.
In addition, recent research \citep{Gottemukkula2019POLYNOMIALFUNCTIONS} has shown that by using normalization to truncate the dynamic range of the pre-activations, the training of Neural Networks with polynomial activations can be stabilized, and converge to reasonable results in simple image classification tasks (MNIST and CIFAR). 
Yet, there are still a few significant challenges in using polynomial activations:
First, to the best of our knowledge, they were not shown to achieve competitive performance (similar to standard activations) on tasks of more realistic scales, such as ImageNet. In addition, the normalization method for dynamic range truncation causes the (truncated) polynomial to increase the signal's bandwidth indefinitely, which is not suitable for aliasing-free CNNs. This normalization was shown to be crucial for convergence even in small tasks and it is reasonable to expect that is even more important for larger tasks. 

\paragraph{Contributions}
In this paper 
\begin{itemize}
    \item We propose the first Alias-Free Convnet (AFC). 
    \item We prove the AFC has both shift-invariant outputs and shift-equivariant internal representations --- even for fractional shifts, where previous models fail.
    \item We show how simple and easy ``black-box'' adversarial attacks built on fractional image translation can degrade a CNN performance, even when the CNN is invariant to integer shifts. In contrast, the AFC has certified robustness to such attacks and superior test accuracy in this regime.
    \item Specifically, the robustness of AFCs is certified for circular shifts and the ideal (Sinc) interpolation kernel. However, we show empirically that AFCs have improved robustness even with other types of translations.
    
    \item Interestingly, our model relies on polynomial activations, and we are the first to demonstrate competitive performance with such activations on ImageNet, to the best of our knowledge.
\end{itemize}
\section{Methods}
\label{sec:methods}
\subsection{Shift invariance and equivariance}
Let $\tau_{\Delta}:L^2(\mathbb{R}^2)\to L^2(\mathbb{R}^2)$ be the translation operator, which shifts a continuous-domain two-dimensional signal by $\Delta\in\mathbb{R}^2$. An operator $f:L^2(\mathbb{R}^2)\to L^2(\mathbb{R}^2)$ is said to be \emph{shift-equivariant} if it commutes with $\tau_{\Delta}$ for every $\Delta$. Namely,
\[
f \left(  \tau_{\Delta} \left( x \right) \right) = \tau_{\Delta} \left( f \left( x \right) \right)
\]
for every $x\in L^2(\mathbb{R}^2)$ and every $\Delta\in\mathbb{R}^2$. \\
An operator $f:L^2(\mathbb{R}^2)\to\mathbb{R}^d$ is said to be \emph{shift-invariant} if its output is invariant to translation of its input, i.e.
\[
f \left(  \tau_{\Delta} \left( x \right) \right) =  f \left( x \right)\,.
\]

The definitions of equivariance and invariance to translations naturally transfer to discrete-domain signals in $L^2(\mathbb{Z}^2)$ and integer shifts $\Delta\in\mathbb{Z}^2$. 
To simplify notations, from now on we will not specify the domain over which operators are defined, and will also omit the subscript $\Delta$ from $\tau$, whenever the meaning is clear from the context.

CNN architectures for classification commonly comprise a \emph{Feature Extractor}, which is mainly composed of convolution layers, and a \emph{Classifier}, which is typically composed of a linear layer and 
a softmax activation.
A sufficient condition for the model to be shift-invariant is that the \emph{Classifier} be shift-invariant, and the \emph{Feature Extractor} be shift-equivariant. 
This is because the composition of a shift-equivariant $f$ and a shift-invariant $g$ yields a shift-invariant function, as
\[
g\left(f\left(\tau\left(x\right)\right)\right)=g\left(\tau\left(f\left(x\right)\right)\right)=g\left(f\left(x\right)\right)\,.
\]
For our discussion, we assume that the \emph{Classifier} is shift-invariant as its inputs are the spatially-averaged channels.
However, the \emph{Feature Extractor} part of CNNs commonly includes also downsampling layers. 
The spatial dimensions of the output of such layers are smaller than the spatial dimensions of their input. 
Therefore, for such layers, shift-equivariance is not a desired property. 
Indeed, when shifting an image by 2 pixels at the input of a layer that performs downsampling by a factor of 2, we expect the output image to shift by only 1 pixel, not 2.
Even worse, when shifting an image by only 1 pixel, it is not clear how precisely the output should shift.
In order to extend the discussion to include these networks, here we consider equivariance w.r.t.~the continuous domain. To simplify the exposition, let us present the definitions for 1D signals, where `discrete' and `continuous' will refer to the signal index we use. 
Namely, a discrete signal $x[n]$ is defined over $n\in\mathbb{Z}$ while a continuous signal $x(t)$ is defined over $t\in\mathbb{R}$.

\begin{definition}[Fractional translation for discrete signals] \label{def:fracShift}
Let  $x[n]$ be a discrete-domain signal and let $\Delta\in\mathbb{R}$ be a (possibly non-integer) shift. 
Then the translation operator $\tau_\Delta$ is defined by $\tau_{\Delta}(x)[n]=z(nT+\Delta)$, where $z(t)$ is the unique $1/2T$-bandlimitted continuous-domain signal satisfying $x[n]=z(nT)$.
\end{definition}

Note that the uniqueness of $z(t)$ in \cref{def:fracShift} is guaranteed by the Nyquist theorem. 
It is also easily verified that this definition does not depend on $T$. 
Equipped with this definition, we can define the following. 

\begin{definition}[shift-equivariance w.r.t.~the cont.~domain]
\label{def:shift-equiv}
An operator $f$ operating on discrete signals is said to be shift-equivariant w.r.t.~the continuous domain if it commutes with fractional shifts. 
Namely, $f (  \tau_{\Delta} ( x ) ) = \tau_{\Delta} ( f ( x ) )$ for every $x\in L^2(\mathbb{Z})$ and every $\Delta\in\mathbb{R}$.

\end{definition}
Similarly, we can define the following.
\begin{definition}[shift-invariance w.r.t.~the cont.~domain]
\label{def:shift-inv}
An operator $f$ operating on discrete signals is said to be shift-invariant w.r.t.~the continuous domain if it is invariant to fractional shifts of its input. 
Namely, $f (  \tau_{\Delta} ( x ) ) = f ( x )$ for every $x\in L^2(\mathbb{Z})$ and every $\Delta\in\mathbb{R}$.
\end{definition}

An important observation is the following.
\begin{proposition} \label{prop:cnn_equivariance_invariance}
In a network comprised of a Feature Extractor and a Classifier, if the Feature Extractor ends with a global average pooling layer, then shift-equivariance w.r.t.~the continuous domain of the Feature Extractor implies shift-invariance  w.r.t.~the continuous domain of the entire model.
\end{proposition}

Indeed, in this case, the Classifier's input is only dependent on the average of the Feature Extractor, which is shift-invariant.
The last statement stems from the fact that when shifting the input of an operator that is shift-equivariant w.r.t.~the continuous domain, the output must be a faithful translated discrete representation of the same continuous signal. 
Namely, there exists some $1/2T$-bandlimited continuous signal $\tilde{f}( t )$  such that $f  (x ) [ n ] = \tilde{f} (nT)$ and  $f ( \tau ( x ) ) [ n ] = \tilde{f} (nT + \Delta ) $. 
Thus, the averages of $f ( x ) [ n]$ and $f ( \tau ( x ) ) [ n ]$ are both equal to the ``DC component'' of $\tilde{f}$, and therefore must be equal.

In order to examine the property of equivariance w.r.t.~continuous domain of CNNs, we shall look at the discrete signal that propagates in a CNN as a representation of a continuous signal, and at each layer as a representation of a continuous operation on the continuous signal.
As shown by \citet{Karras2021Alias-FreeNetworks}, aliasing in the discrete representation prevents shift-invariance of CNNs since it decouples the discrete signal from its continuous equivalent. 
In contrast, they have shown that alias-free operations preserve shift-equivariance w.r.t.~continuous domain, and lead to shift-invariant CNNs.
There, \citet{Karras2021Alias-FreeNetworks} have shown that convolutions and downsamplers which are properly treated using low-pass filters (LPFs), are indeed alias-free and thus shift-equivariant w.r.t.~the continuous domain. 
In addition, they proposed a method to reduce the implicit aliasing of non-linearities which we describe next.

In the continuous domain, pointwise non-linearities may induce indefinitely high new frequencies. 
Applying a pointwise non-linearity in the discrete domain is equivalent to sampling a continuous signal after applying the pointwise non-linearity --- which may break the Nyquist condition and cause aliasing. This implies that pointwise nonlinearities applied in the discrete domain are generally not shift-invariant w.r.t.~the continuous domain. 
Using upsampling before the non-linearity may solve this problem since it increases the frequency support that does not cause aliasing.
However, this approach cannot generally prevent aliasing, since the new frequencies generated by non-linear operations can be arbitrarily high. 
For example, the outputs of non-differentiable operations such as ReLU can have infinite support in the frequency domain, thus aliasing will be induced for every finite upsampling factor.

In this study, we propose replacing non-linear operations with a band-limited preserving alternative --- polynomial functions. 
The proposed scheme for an aliasing-free polynomial function of degree $d$ is defined in \Cref{algo:af-poly}.
In this algorithm, $\mathrm{Upsample}_{z}$ performs upsampling by a factor $z$ (i.e.~resampling the input continuous signal at a $z\times$ larger sampling frequency), $\mathrm{LPF}_{z}$ is an ideal low-pass filter with cut-off $z$, $\mathrm{Downsample}_{z}$ performs downsampling by a factor $z$ (i.e.~dividing the sample frequency by $z$), and 
\begin{equation}
    \mathrm{Poly}_d(x)=\sum_{i=0}^d a_ix^i \,.
\end{equation}


\begin{algorithm}[h!]
    \caption{Alias-free polynomial activation}
    \label{algo:af-poly}
\begin{algorithmic}
    \item {\bfseries Input:} $x$ - input signal, $\mathrm{Poly}_{d}$ - polynomial of degree $d$.
    \item $x_{\mathrm{up}} \gets \mathrm{Upsample}_{\frac{d+1}{2}}\left(x\right)$
    \item $y_{\mathrm{poly}} \gets \mathrm{Poly}_{d} \left( x_{\mathrm{up}} \right)$ 
    \item $y_{\mathrm{LPF}} \gets \mathrm{LPF}_{\frac{2}{d+1}}\left(y_{\mathrm{poly}}\right)$
    \item $y \gets \mathrm{Downsample}_{\frac{d+1}{2}}\left(y_{{\mathrm{LPF}}}\right)$
    \item {\bfseries Output:} $y$
\end{algorithmic}
\end{algorithm}

\begin{figure}[ht]
   \centering
   \includegraphics[width=0.98\linewidth]{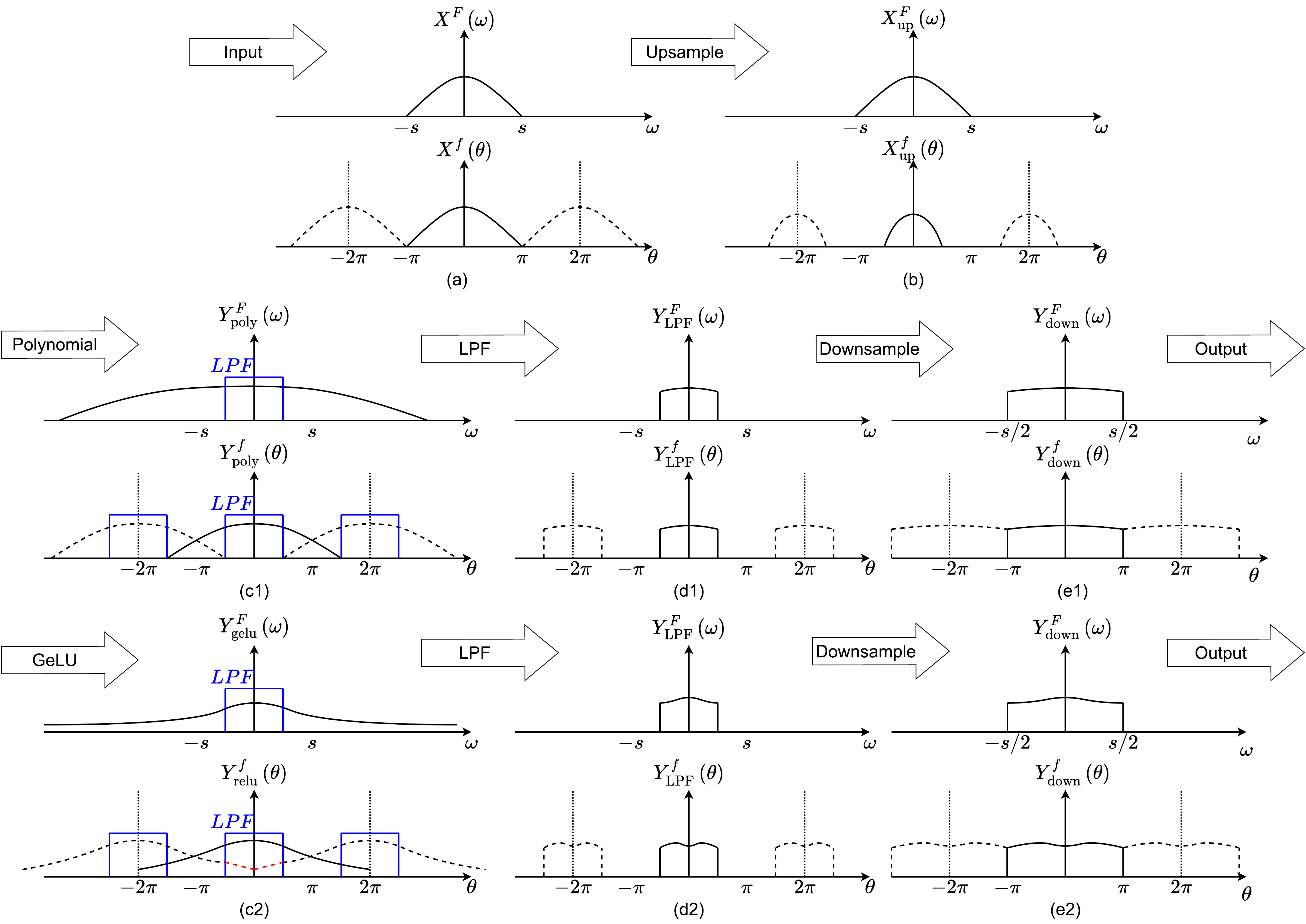}

   \caption{\textbf{A demonstration of the proposed non-linearities in the frequency domain.} The top plot at each panel represents the signal in the continuous domain, and the bottom represents the discrete domain.
   Where the input (a) is upsampled it shrinks its frequency response, expanding the allowed frequencies (b).
   Applying the polynomial activation expands the frequency response support by as factor $d$, without causing aliasing in the relevant frequencies (c.1). Thus, the discrete signal remains a faithful representation of the continuous signal after applying LPF (d1) and downsample back to the same spatial size (d2). 
   However, applying GeLU expands the support infinitely (c.2). This leads to an aliasing effect --- interference in the relevant frequencies marked in red in (c2). This causes the discrete signal not to be a correct representation of the continuous one, after LPF (d2) and downsampling (e2).}
   \label{fig:poly-alias}   
  
\end{figure}

The practical implementations of the operations above are described in \Cref{sec:implementation} and \Cref{sec:implementation-appendix}.
Our contribution to the general framework that has been presented by \citet{Karras2021Alias-FreeNetworks} is the usage of polynomial activations, which extends the frequency bandwidth in a limited fashion, unlike other non-linearities. 
Hence, by using appropriate upsampling as in \Cref{algo:af-poly}, aliasing can be avoided, as described in \Cref{fig:poly-alias}. 
Specifically, in \Cref{sec:shift-invariance-proof} we prove  the following.

\begin{proposition}\label{prop:af-poly-invariance}
    The operator defined by \Cref{algo:af-poly} is shift-equivariant w.r.t.~the continuous domain.
\end{proposition}

By combining \cref{prop:cnn_equivariance_invariance} and \cref{prop:af-poly-invariance} with the shift-equivariance the other layers (as described above), we conclude the network output is shift-invariant. 
Next, we describe the proposed process of non-linearities in the frequency domain, which is additionally demonstrated in \Cref{fig:poly-alias}.
In the first step, the input $x$ is upsampled, leading to a contraction of its support in the frequency domain (\cref{fig:poly-alias}{(b)}). 
Effectively, it expands the range of allowed new frequencies generated by the following non-linearity (\cref{fig:poly-alias}{(c1)}). Then, a low-pass filter is applied in order to prevent aliasing in the following downsampling layer (\cref{fig:poly-alias}{(d1), (e1)}).
Overall, for an upsampling factor that is appropriate for the frequency expansion of the polynomial, the effective frequencies for the output are not being overlapped at any of the steps, thus aliasing is prevented.
However, in the case of non-linearities that do not preserve the band-limited property, upsampling cannot prevent the frequency overlap in (\cref{fig:poly-alias}(c2)).

\begin{figure}[ht]
    \centering
    \includegraphics[width=0.75\linewidth]{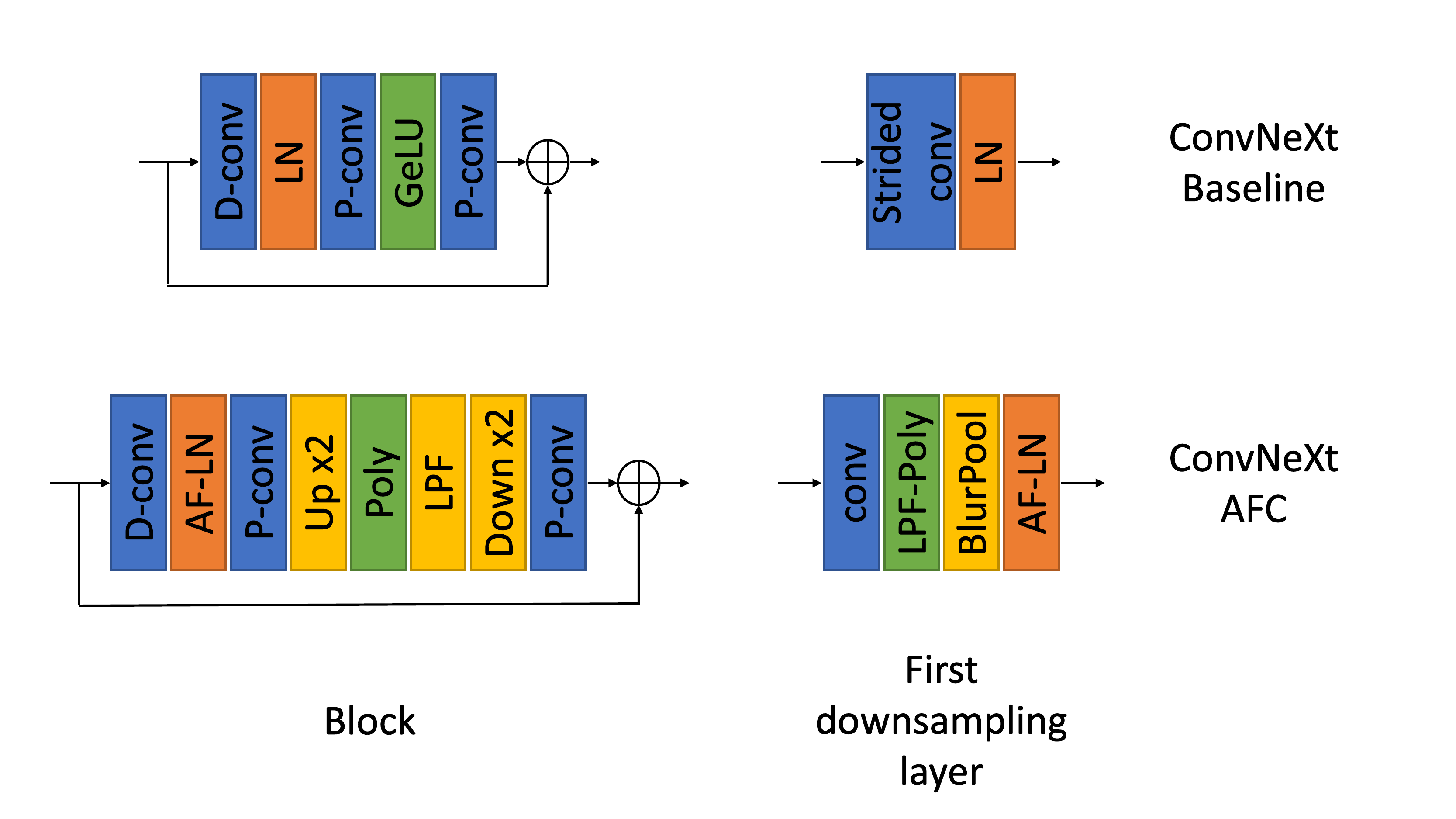}
    
    \caption{\textbf{ConvNeXt baseline architecture vs AFC modifications.}
    D-conv: depthwise convolution $7 \times 7$, P-Conv: pointwise convolution, Strided-conv: convolution $4 \times 4$, stride $4$.
    LN: Layer Norm, AF-LN: Alias free Layer Norm, Poly: Polynomial activation. Up x2: Upsample x2, LPF: ideal LPF with cutoff 0.5, Down x2: Downsample x2. Detailed explanations about BlurPool, Poly and LPF-Poly activations can be found in \Cref{sec:implementation}.}
    \label{fig:model-modifications}    
\end{figure}

\section{Implementation}
\label{sec:implementation}
We propose an Alias-Free Convnet (AFC), based on the ConvNeXt architecture \citep{Liu2022A2020s}, which has been shown to achieve state-of-the-art results in image classification tasks. 
We modify the layers which suffer from aliasing (as described in \cref{fig:model-modifications}) so that the convnet is completely free of aliasing. 
The theoretical derivation in \Cref{sec:methods} assumes infinite-length discrete signals, hence cannot be directly applied in practical systems. 
However, it can be naturally used by limiting the discussion to circular translations, which implies that the continuous signals are periodic. 
In this case, the theoretical results from \Cref{sec:methods} can be equivalently attained with finite-length signals using our following implementation.

\paragraph{Convolution}
We use circular convolutions to meet the periodic signal assumption, as described above.
This is practically done by replacing zero padding with circular padding, similarly to \citet{Chaman2020TrulyNetworks}.

\paragraph{BlurPool}
Similarly to the model presented by \citet{Zhang2019MakingAgain}, we separate strided convolutions into linear convolution and downsampling operations. 
The downsampling operation is replaced by BlurPool, which applies sub-sampling after low-pass filtering. 
Instead of implementing a low-pass filter using convolutions with custom fixed kernels, we implement an ``ideal low-pass filter'' by truncating high frequencies in the Fourier domain. 
Specifically, we transform the input to the Fourier domain using Pytorch FFT kernel \citep{NEURIPS2019_9015}, zero out the relevant frequencies, and transform it back to the spatial domain. 
This is an efficient implementation of downsampling after applying multiplication with filter $H^{2D}$ in DFT domain, which is defined as 
\begin{equation}
H^{2D} = HH^{T} \,,
\end{equation}
where for stride $s$ and spatial-domain size $N \times N$ , $H$ is defined as
\begin{equation}
H[k]=\begin{cases}
1, & 0\leq k<\frac{N}{2s}\,, \\
0, & \frac{N}{2s}\leq k\leq\frac{3N}{2s}\,,  \\
1, & \frac{3N}{2s}<k\leq N-1\,.
\end{cases} 
\end{equation}
A derivation of this filter can be found in \Cref{sec:implementation-appendix}.

\paragraph{Activation function}
We replace the original GeLU activation with a polynomial function of degree 2, whose coefficients are trainable parameters, per channel:
\begin{equation}
    \mathrm{Poly}_2(x)=a_0+a_1x+a_2x^2 \,.
\end{equation}
The coefficients $\{a_0,a_1,a_2\}$ are initialized by fitting this function to the GeLU, as proposed by \citet{Gottemukkula2019POLYNOMIALFUNCTIONS}. 
All activation functions are wrapped according to the alias-free technique presented in \Cref{algo:af-poly}.
Generally, replacing the activation function in a Deep Neural Network may change the scale of the propagated activation, thus requiring adjusting the weight initialization. 
In our experiments the activation scale had a large impact on the achieved accuracy, thus searching for an appropriate scale factor was required.
Details regarding the activation tuning can be found in \cref{sec:poly-append}.
Overall, in our case (polynomial activation in ConvNeXt) using the appropriate scale seemed to recover most or all of the lost accuracy.

\paragraph{Normalization}
ConvNeXt model implementation uses a variation of LayerNorm, which centers and scales each pixel according to its mean and standard deviation over channels, respectively. 
The scaling operation requires the multiplication of each pixel with a different scalar which, like other point-wise non-linearities, is not alias-free. 
We construct an alias-free alternative by using scaling per layer instead of scaling per pixel, i.e.~all pixels are scaled by the standard deviation of the layer, which is shift-equivariant w.r.t.~the continuous domain.
Although eliminating aliasing effects, this modification caused a small reduction in the model accuracy, as we shall see later. 
We hypothesize this reduction results from the ``normalization-per-pixel'' operation functioning as an additional non-linearity, which enlarges the model capacity. 
Yet, this modification is required for the property of shit-equivariance w.r.t. continuous domain, which leads to an overall improvement in terms of robustness to sub-pixel image translations, as shown in \Cref{sec:experiments}.


\paragraph{First downsample layer}
Unlike other CNN architectures that were examined in the context of aliasing prevention, ConvNeXt does not have a non-linearity before the first downsampling layer. 
Thus, due to the commutativity of convolutions with the LPFs, we cannot replace the first downsampling operation with BlurPool --- since this is equivalent to applying a low-pass filter directly on the input, effectively reducing its resolution. 
Such composition may prevent the model from using high-frequency features and lead to a reduction in the model's accuracy.
To solve this problem, we add an additional activation function before the first BlurPool. 
For computation efficiency, instead of using the regular scheme which requires upsampling before the activation, we replace the usual activation  $\mathrm{Poly}_2(x)$ with
\begin{equation}
    \mathrm{LPFPoly_2}(x)= a_0 + a_1 x + a_2 x \cdot \mathrm{LPF}_{\frac{3}{4}}\left(x \right) \,.
\end{equation}
This modification of the polynomial activation leads to a smaller increase of the signal bandwidth. 
Thus, it does not require upsampling to avoid aliasing when it is followed by an LPF, as in the first BlurPool.
Specifically, since it is followed by a BlurPool with a cutoff $1/4$, The maximally allowed cutoff for the LPF-Poly's filter is $3/4$.
More details on this activation function can be found in \Cref{sec:lpf-poly-append}.

\section{Experiments} \label{sec:experiments}
We compare our Alias-Free Convnet (AFC) model to the baseline ConvNeXt model and to the previous integer shift-invariant method Adaptive Polyphase Sampling (APS) \citep{Chaman2020TrulyNetworks}. 
We implemented all models with cyclic convolutions and trained them on ImageNet \citep{Deng2009ImageNet:Database} according to the ConvNeXt training regime \citep{Liu2022A2020s}. 
The experiments were conducted with circular translations similarly to the setting in previous works \citep{Zhang2019MakingAgain, Chaman2020TrulyNetworks}.
For sub-pixel translations, we used our ``ideal upsampling'' implementation \Cref{algo:Up-sample} (i.e., translation by $m/n$ pixels was conducted by upsampling by $n$, translating by $m$ pixels and downsampling by $n$).

\subsection{Shift equivariance}
\begin{figure}[ht]
    \centering
    \includegraphics[width=0.45\linewidth]{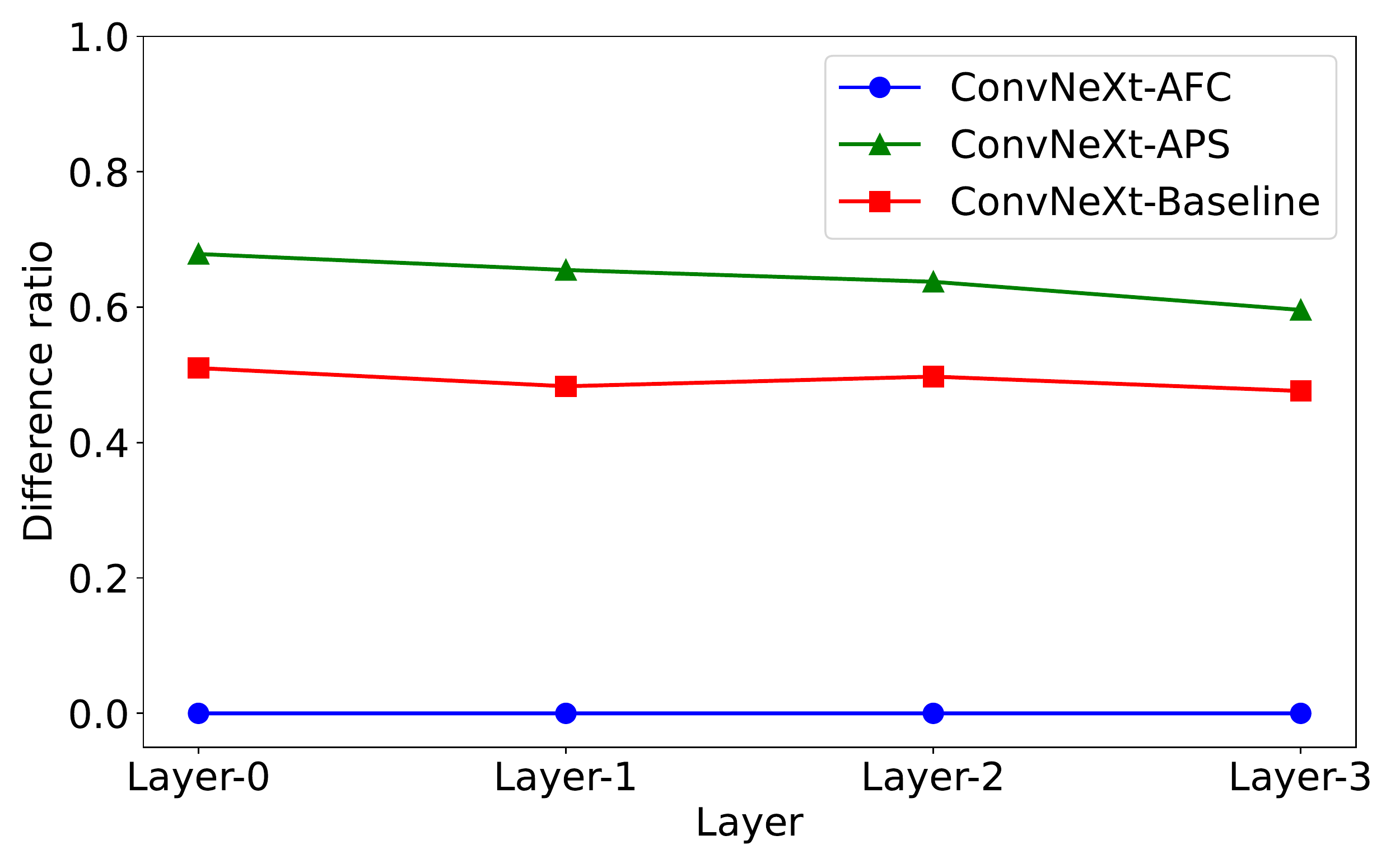}
   \caption{\textbf{Shift-equivariance measure w.r.t. continuous signal.}
   The averaged difference (\cref{eq:diff}) for $1/2$ pixel translated inputs (y-axis), across all layers (x-axis).
   This experiment was run on 64 random samples from the validation set.
    While the AFC model has practically 0 difference, the baseline and APS models have at least 50\% difference across all layers.}
   \label{fig:equivarance}
\end{figure}

Our model is designed to be not only shift-invariant (in terms of classification output), but also to have a Feature-Extractor that is shift-equivariant w.r.t.~to the continuous domain. 
We verified this property by examining the response of the output of each of the layers to a translation of $\frac{1}{2}$ pixel in the input image. 
This was done by propagating the two translated inputs and measuring the difference between their outputs in each layer, after upsampling back to the input's spatial size. 
The results in \Cref{fig:equivarance} show, in each layer, the normalized difference between the two translated layer outputs $y^0$ and $y^1$, after they were averaged across all $HW$ pixels (indexed by $i,j$) and $C$ channels (indexed by~$c$),
\begin{equation} \label{eq:diff}
    \mathrm{diff} \triangleq \frac{1}{CHW}\sum_{c,i,j} \frac{ \left| y^{0}_{c,i,j} - y^{1}_{c,i,j} \right|}{\max \left( \left| y^{0}_{c,i,j} \right|, \left| y^{1}_{c,i,j} \right|\right) + \varepsilon}\, ,
\end{equation}
where $\varepsilon=10^{-9}$ was added in the denominator to avoid division by $0$.
The results show that ConvNeXt-AFC has only a negligible difference in the continuous representation of the translated responses at each layer, e.g.~$y^0 = y^1$, which means it is indeed shift-equivariant w.r.t.~the continuous domain (up to numerical error). 
In contrast, in the case of the baseline and APS models, the upsampled signals differ by more than 50\% across all the layers.

\subsection{Consistency and Classification accuracy}
The main measure used so far to quantify the shift-invariance of a model is called ``consistency'' \citep{Azulay2019WhyTransformations,Chaman2020TrulyNetworks}, which is the percentage of predictions changed on the test set following an image shift. 
Previously this measure has been used with integer shift values, however, in \Cref{table:aal-modifcations-accuracy} we test it also under sub-pixel shifts. 
We see that the changes we add to the baseline model gradually improve its consistency, until we reach 100\%.
In contrast, the previous APS approach \citep{Chaman2020TrulyNetworks} is near 100\% consistent to integer shifts (due to numerical accuracy), but for fractional shifts, it only has slightly higher consistency than the baseline.
Even though the alias-free modifications in our model lead to perfect consistency, they cause a 1.08\% reduction in the (standard) test accuracy. 
We conclude that the main source of accuracy reduction is the modification of the normalization layer, as explained in \Cref{sec:implementation}. 
However, as we shall see next, despite such a reduction in accuracy, our model outperforms the previous models in adversarial shifts setting, due to its increased robustness.

\begin{table}[t]
\caption{\textbf{Alias-free modifications ImageNet accuracy and shift-consistency effect.}
Integer shift consistency is defined as the percentage of test samples that did not change their prediction following a random integer translation.
Fractional shift consistency is defined as the percentage of test samples that did not change their prediction following a random half-pixel translation. 
Consistency was averaged on five runs on ImageNet validation set with random seeds.
The final AFC model is 100\% consistent to both integer and fractional translations.
Note that though the APS  model \citep{Chaman2020TrulyNetworks} exhibits near 100\% integer shifts consistency (as expected), it has only slightly better consistency than the baseline model in terms of fractional shift consistency.}
  \label{table:aal-modifcations-accuracy}
\small
\centering
\resizebox{1.0\linewidth}{!}{
\begin{tabular}{lcccccc}
\toprule
Model modification                  & \specialcell{Test\\accuracy} & Change & \specialcell{Integer\\shift consistency} & Change & \specialcell{Fractional \\shift consistency} & Change \\
\midrule
\specialcellleft{ConvNeXt-Baseline\\ \citep{Liu2022A2020s}}                    & 82.12         &        & 94.816                    &        & 92.034                    &        \\
+ Polynomial activation           & 81.77         & -0.35  & 95.126                    & 0.31   & 92.708                    & 0.67   \\
+ BlurPool                        & 78.99         & -3.12  & 96.635                    & 1.82   & 96.572                    & 4.54   \\
+ First layer activation          & 81.51         & -0.61  & 97.354                    & 2.54   & 97.347                    & 5.31   \\
+ AF LayerNorm                    & 80.66         & -1.46  & 97.030                    & 2.21   & 96.990                    & 4.96   \\
\specialcellleft{+ Activation upsample\\(ConvNeXt-AFC, ours)} & 81.04         & -1.08  & 100.000                   & 5.18   & 100.000                   & 7.97   \\
\midrule
\specialcellleft{ConvNeXt-APS\\ \citep{Chaman2020TrulyNetworks}}                     & 82.11         & -0.01  & 99.998                   & 5.18   & 93.227                    & 1.19  \\
\bottomrule
\end{tabular}
}
\end{table}

\begin{table}[t]
\caption{\textbf{Translation adversarial accuracy (ImageNet). }
Left column: Test accuracy. 
Right columns: adversarial accuracy defined as the percentage of correctly classified samples for each translation in the corresponding set: \cref{eq: T1}, \cref{eq: T2} or \cref{eq: T3} with $k=12$.}
\label{table:translation-robustness}
\small
\centering
\begin{tabular}{lcccccc}
\toprule

model & \specialcell{Test} &  \specialcell{Adversarial\\integer grid} &  \specialcell{Adversarial\\ half-pixel grid } & \specialcell{Adversarial \\fractional grid}\\
\midrule
\specialcellleft{ConvNeXt-Baseline\\ \citep{Liu2022A2020s}}   & 82.12        &  76.63             & 73.65     & 77.82    \\
\specialcellleft{ConvNeXt-APS\\  \citep{Chaman2020TrulyNetworks}}      & 82.11        & 82.11              & 79.68     & 76.31    \\
ConvNeXt-AFC  (ours)       & 81.04          & 81.04             & 81.04       & 81.04     \\
\bottomrule
                           
\end{tabular}
\end{table}

\subsection{Translation robustness}
Since standard models are not invariant to image translation, this might be exploited as a very easy form of a ``black-box'' adversarial attack: we simply move the image until we notice the prediction is changed.  
We examine this vulnerability, to assess each model's actual robustness to translations. 
For each sample, we performed all possible translations in some set $T$, and checked the resulting classification for each shift. 
We define the adversarial accuracy corresponding to $T$ as the portion of samples that are classified correctly for all translations in $T$.
We tested three types of basic translation grids --- Integer, Half pixel and Fractional:
    \begin{equation} \label{eq: T1}
              T_\mathrm{integer} = \left\{ \left( i,\ j \right)\,|\, 1 \leq i,j \leq 31 \right\}   
    \end{equation}
    \begin{equation} \label{eq: T2}
                T_\mathrm{half} = \left\{ \left( \frac{i}{2},\ \frac{j}{2}\right) \,|\, 1 \leq i,j \leq 63 \right\}
    \end{equation}
    \begin{equation} \label{eq: T3}        
        T_{\mathrm{frac}, k} = \left\{ \left(\frac{m_{1}}{n_{1}},\frac{m_{2}}{n_{2}}\right)\,|\, 1\leq m_{1,2}\leq n_{1,2}\leq k \right\} 
    \end{equation}  
In \Cref{table:translation-robustness} we observe the adversarial robustness with respect to these translation sets. 
In the baseline model the test accuracy of 82.1\% drops to 76.63\% for integer grid and to 73.65\% for half-pixel grid accuracy. 
This significant drop reflects that more than 10\% of the correctly classified test set samples may be misclassified due to translations.
The APS model \citep{Chaman2020TrulyNetworks} is, by construction, robust to integer translations and therefore has no accuracy reduction in the integer grid. 
However, it gets even worse results than the baseline in fractional adversarial accuracy (76.31\% vs 77.82\%).
In contrast, our AFC model is invariant to any of these shifts, and therefore its accuracy remains constant at 81.03\%, surpassing the other models. 
This robustness is `certified', and will not be compromised with larger translation sets, or other types of attacks (e.g., white box attacks) which can potentially decrease the performance of the other models even more.

\subsection{Out-of-distribution robustness}
\begin{figure}[ht]
    \begin{center}
    \begin{subfigure}[b]{0.45\linewidth}
    \centering
    \includegraphics[width=\linewidth]{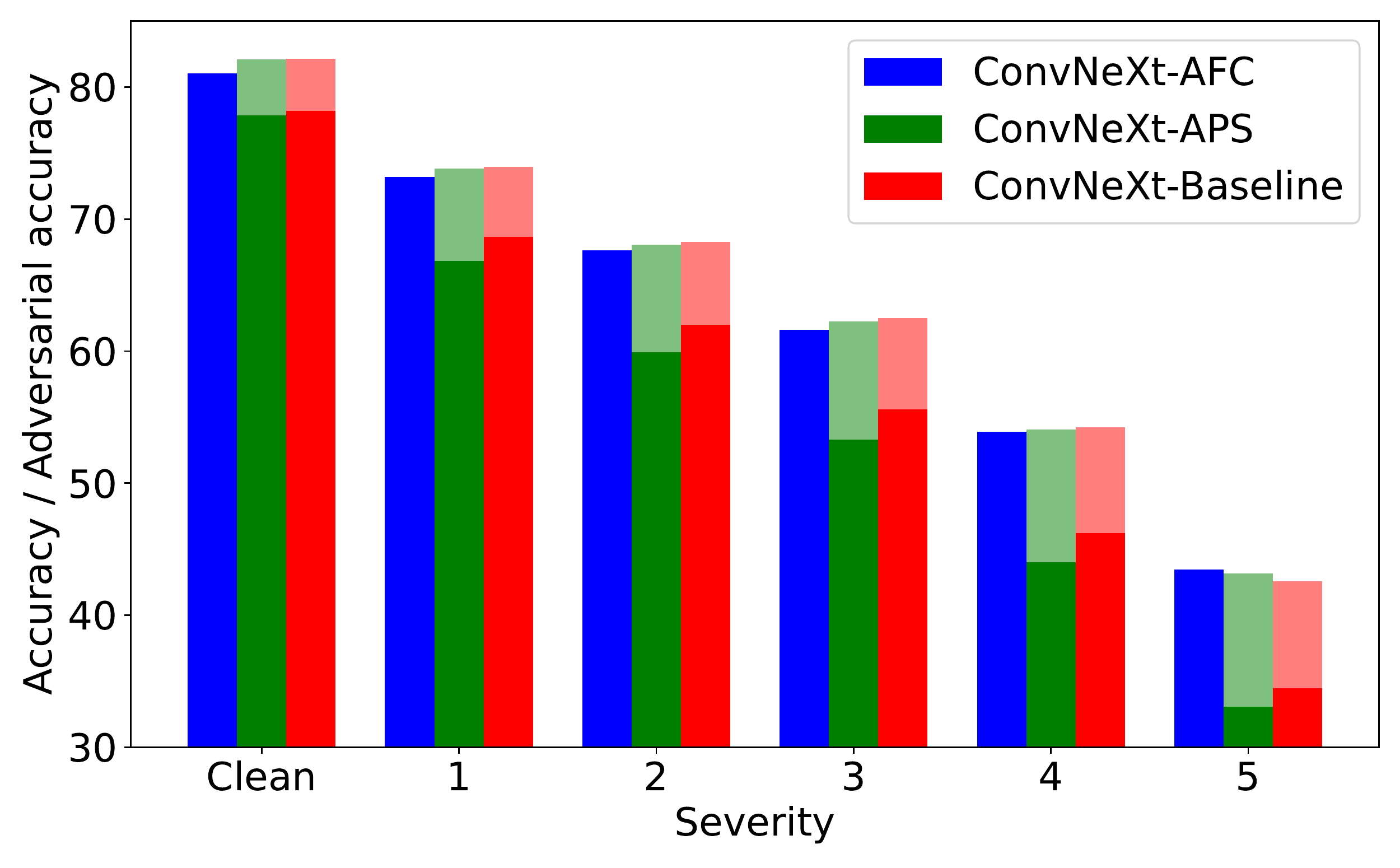}
    \end{subfigure}
    \begin{subfigure}[b]{0.45\linewidth}
    \centering
    \includegraphics[width=\linewidth]{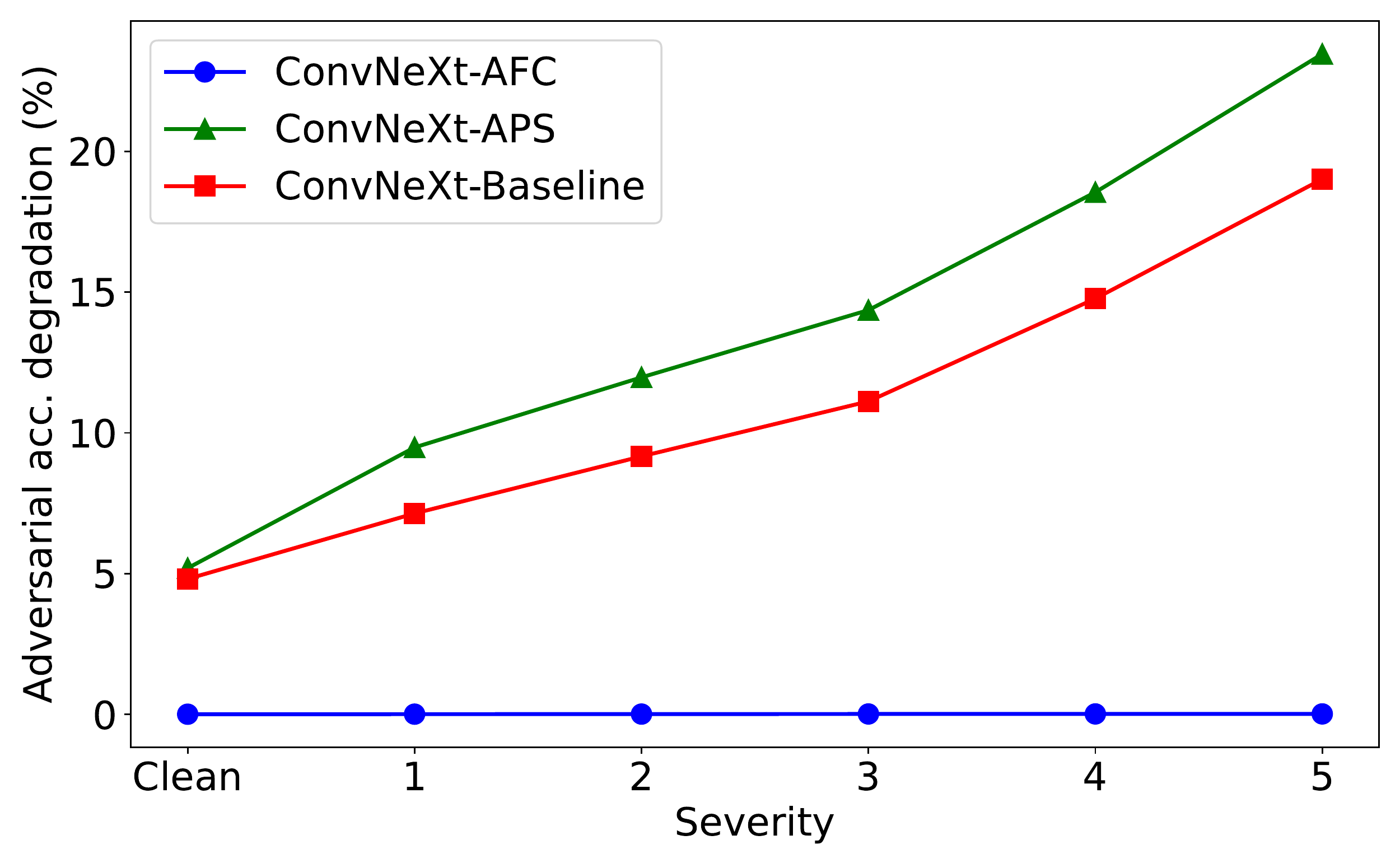}
    \end{subfigure}
   \caption{ \textbf{Adversarial accuracy with image corruptions.}
    \textbf{Left}: ImageNet-C accuracy (solid) vs. adversarial fractional grid accuracy (transparent). 
    \textbf{Right}: Accuracy vs. adversarial accuracy difference (percentage). 
    ConvNeXt-AFC (ours) ImageNet-C accuracy is not affected by translations, while in ConvNeXt-APS and ConvNeXt-Basline the relative accuracy degradation as a result of translations increases with the corruption severity.
  }
\label{fig:imagenet_c}

   \end{center}
   \vskip -0.2in
\end{figure}

While in our model shift-invariance is guaranteed, it is merely a learned property in other models, and thus may only be partially generalized to out-of-distribution images \citep{Azulay2019WhyTransformations}. 
To evaluate this hypothesis, we measured robustness to fractional translations on ImageNet-C \citep{Hendrycks2019BenchmarkingPerturbations}, which contains common corruptions of ImageNet images in ascending severity levels. 
We used fractional grid attack with a minimal translation of $1/7$ a pixel (\cref{eq: T3} with $k=7$).
The results are visualized in \Cref{fig:imagenet_c}. 
As our model's robustness to translations is guaranteed, it has no accuracy reduction caused by translations. 
In contrast, the other models' vulnerability to translation attacks increases with the severity of the corruption; ConvneXt-AFC relative accuracy degradation due to fractional translations increases from 5\% to as much as 23\% in the highest corruption severity. 
This indicates that the generalization of learned shift-invariance is limited in comparison to architectural shift-invariance.


\subsection{Robustness to other shifts}
We next test the models' robustness to other types of translations, where our model's shift-invariance guarantee conditions are not satisfied.
\subsubsection{Zero-padding, bilinear-interpolation}
\begin{figure}[ht]
    \centering
    \begin{subfigure}[b]{0.32\linewidth}
    \centering
    \includegraphics[width=\linewidth]{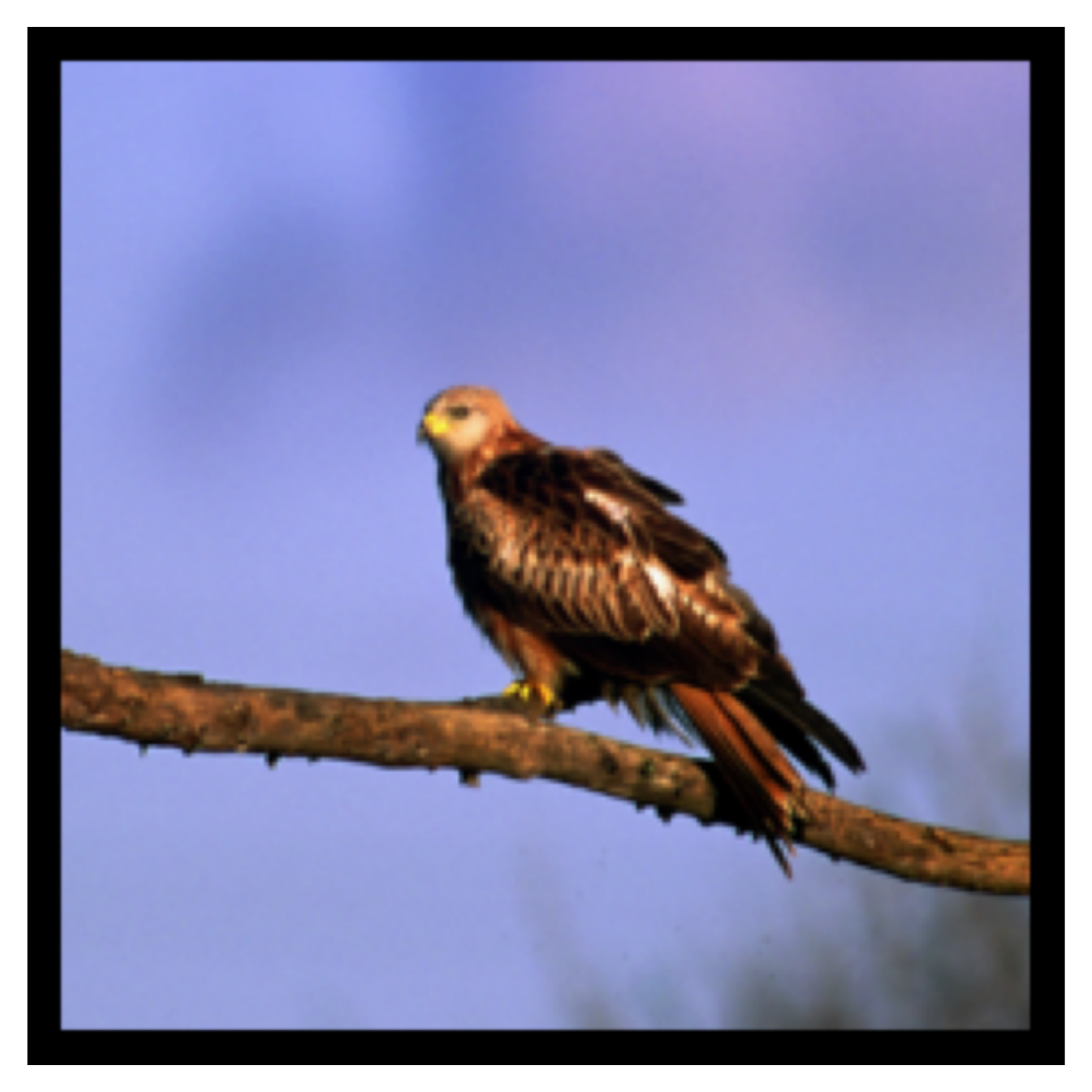}
    \caption{Original image}
    \end{subfigure}
    \begin{subfigure}[b]{0.32\linewidth}
    \centering
    \includegraphics[width=\linewidth]{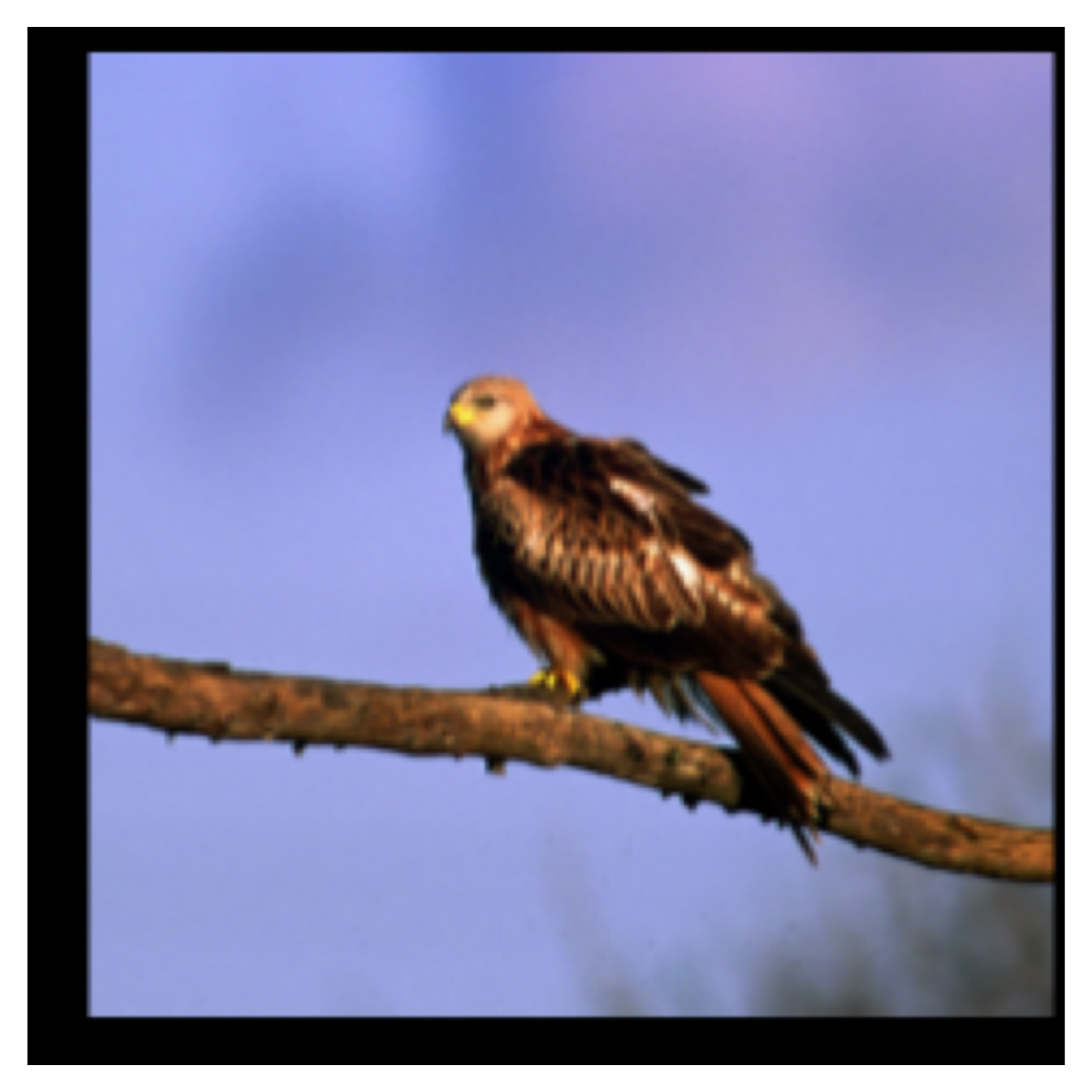}
    \caption{Shifted image}
    \end{subfigure}
\caption{
\textbf{  Visualization of shift attacks similar to the framework of \citet{Engstrom2017ExploringRobustness}.}
  (a) The original image is zero-padded in 8 pixels in each direction. 
  The attack is a translation of up to 8 pixels in each direction, e.g.~(b) is a translation of $6$ and $-2.5$ pixels in $x$ and $y$ axes respectively. Sub-pixel translations are done using bilinear interpolation.
  }
\label{fig:zero_pad_shifts}
\end{figure}

We tested the models' robustness to translation using the framework presented by \citet{Engstrom2017ExploringRobustness}, originally designed to test the robustness of classification models to translations and rotations.
We zero-pad the images by $8$ pixels and translate by (a possibly fractional) amount limited by $8$ pixels, so there are no artifacts due to circular translations, nor data loss. The remaining parts are zero-padded and fractional translations are done using bilinear interpolation (see \Cref{fig:zero_pad_shifts}).
The results in \Cref{fig:crop-shift-bilinear} (left) show the models' adversarial accuracy to this attack with different grid sizes.
Although our model is not perfectly invariant to the performed translations due to the bilinear interpolation, it outperforms the other models by more than 4\% at the largest tested grid.

\subsubsection{Crop-shift}
\begin{figure}[ht]
    \centering
    \begin{subfigure}[b]{0.32\linewidth}
    \centering
    \includegraphics[width=\linewidth]{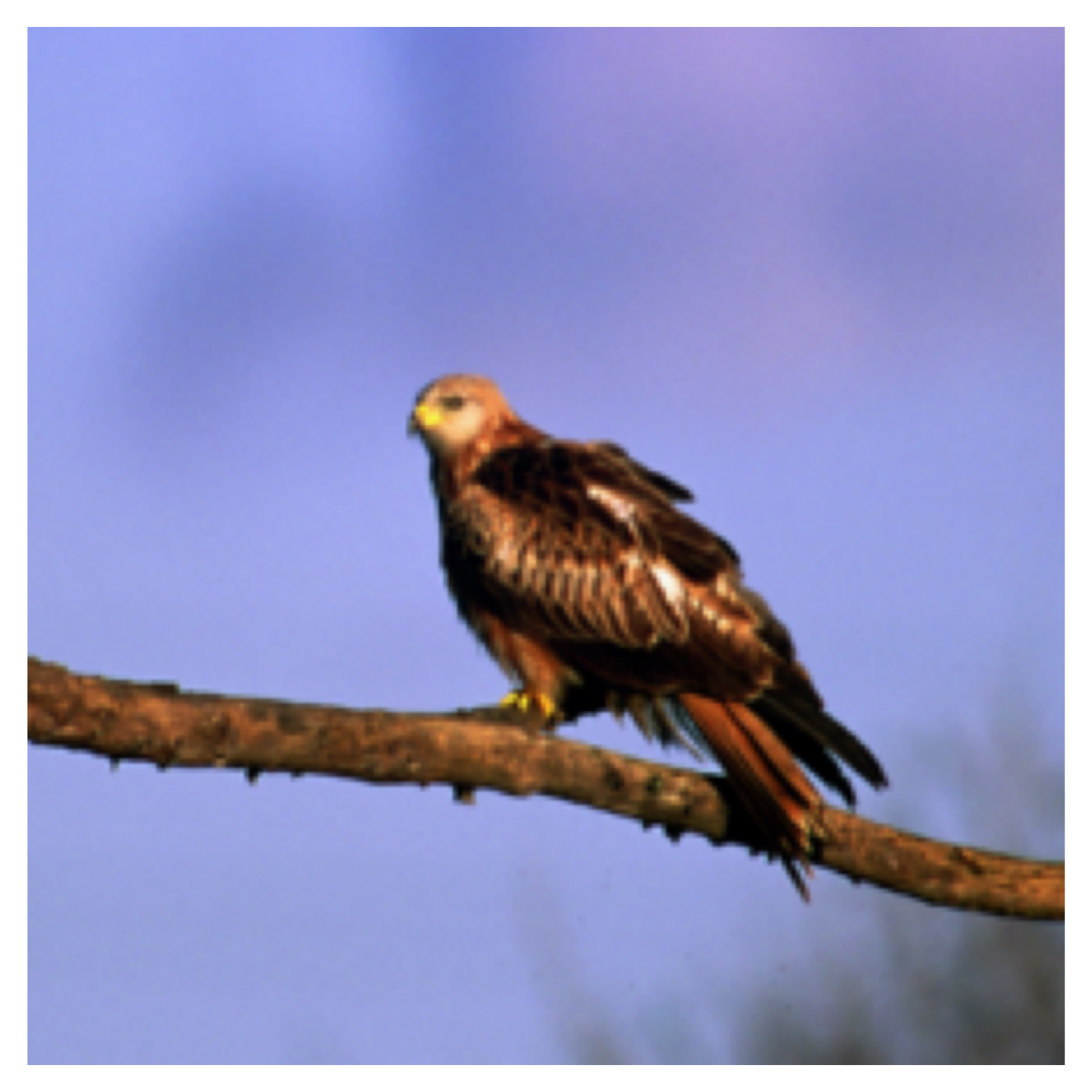}
    \caption{Original image}
    \label{fig:circular_crop_shifts_a}
    \end{subfigure}
    \begin{subfigure}[b]{0.32\linewidth}
    \centering
    \includegraphics[width=\linewidth]{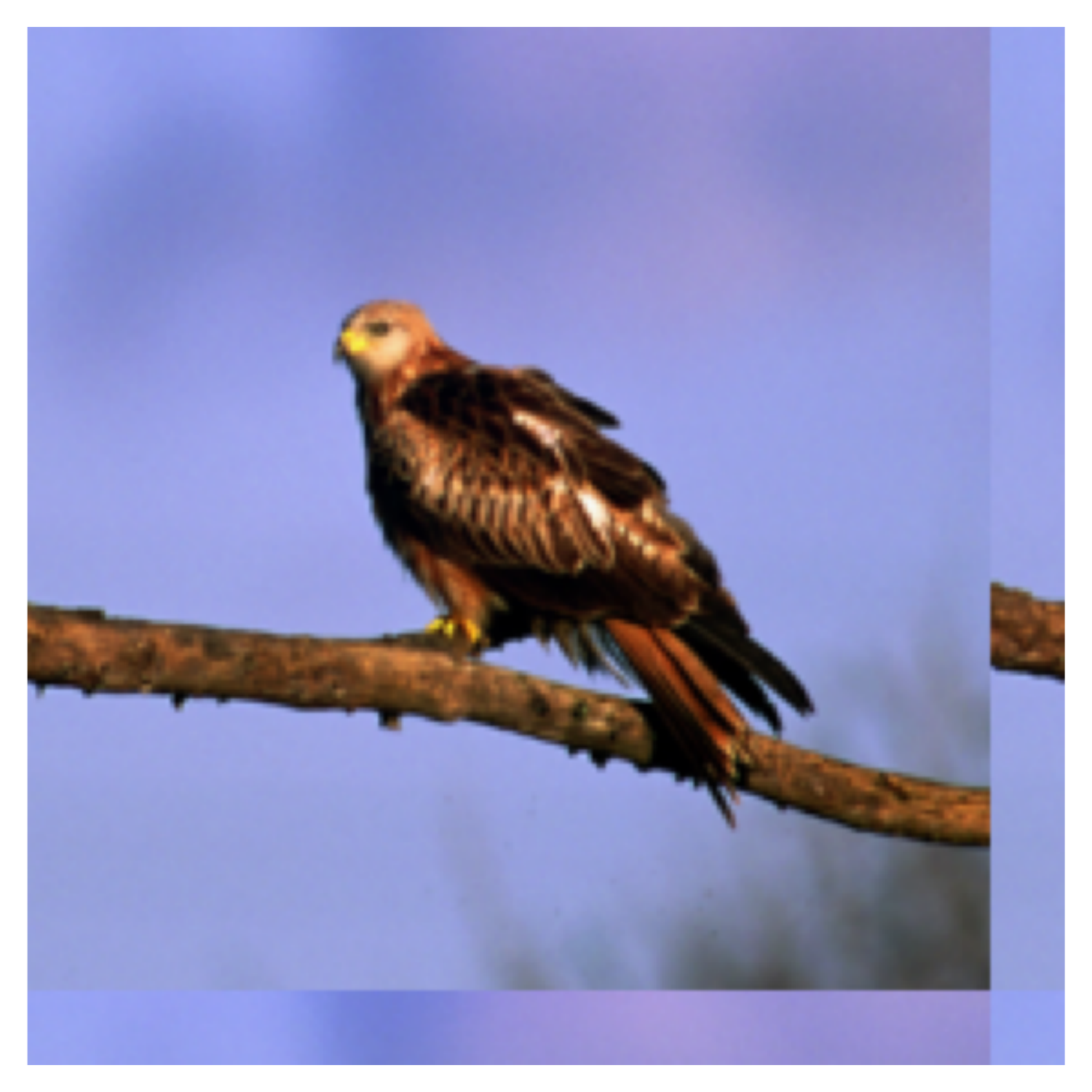}
    \caption{Circular shifted image}
    \end{subfigure}
    \begin{subfigure}[b]{0.32\linewidth}
    \centering
    \includegraphics[width=\linewidth]{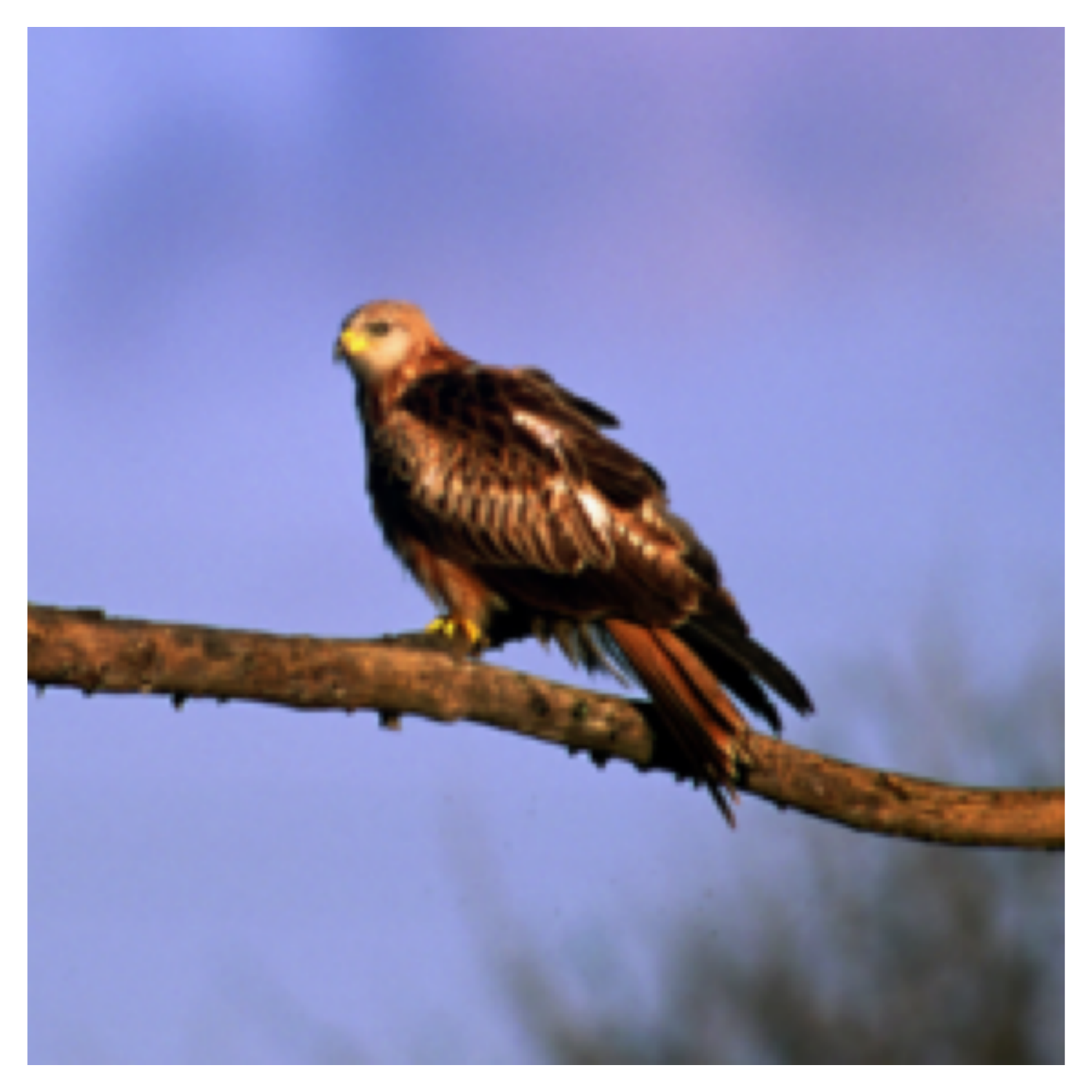}
    \caption{Crop-shifted image}
    \end{subfigure}
\caption{\textbf{Visualization of used attacks.} 
(a) Original ImageNet \citep{Deng2009ImageNet:Database} validation-set image  --- 224 × 224 center crop of the
original 256 × 256 image. (b): Circular shift of 16 pixels in $x$ and $y$ axes. 
(c): ``Crop-shift'' of the original image of 16 pixels in $x$ and $y$ axes; the cropped area is shifted, modeling moving the camera with respect to the scene in the bottom-right direction. 
The top-left part of the circular shifted and crop-shifted images are equal to the bottom-right part of the original image.
The bottom and right edges of the circular shifted image consist of the top and left edges of the original image, causing unrealistic artifacts, while in the crop-shifted change we change the information from the scene.
}
\label{fig:circular_crop_shifts}
\end{figure}

\begin{figure}[ht]
    \begin{center}
    \begin{subfigure}[b]{0.45\linewidth}
    \centering
    \includegraphics[width=\linewidth]{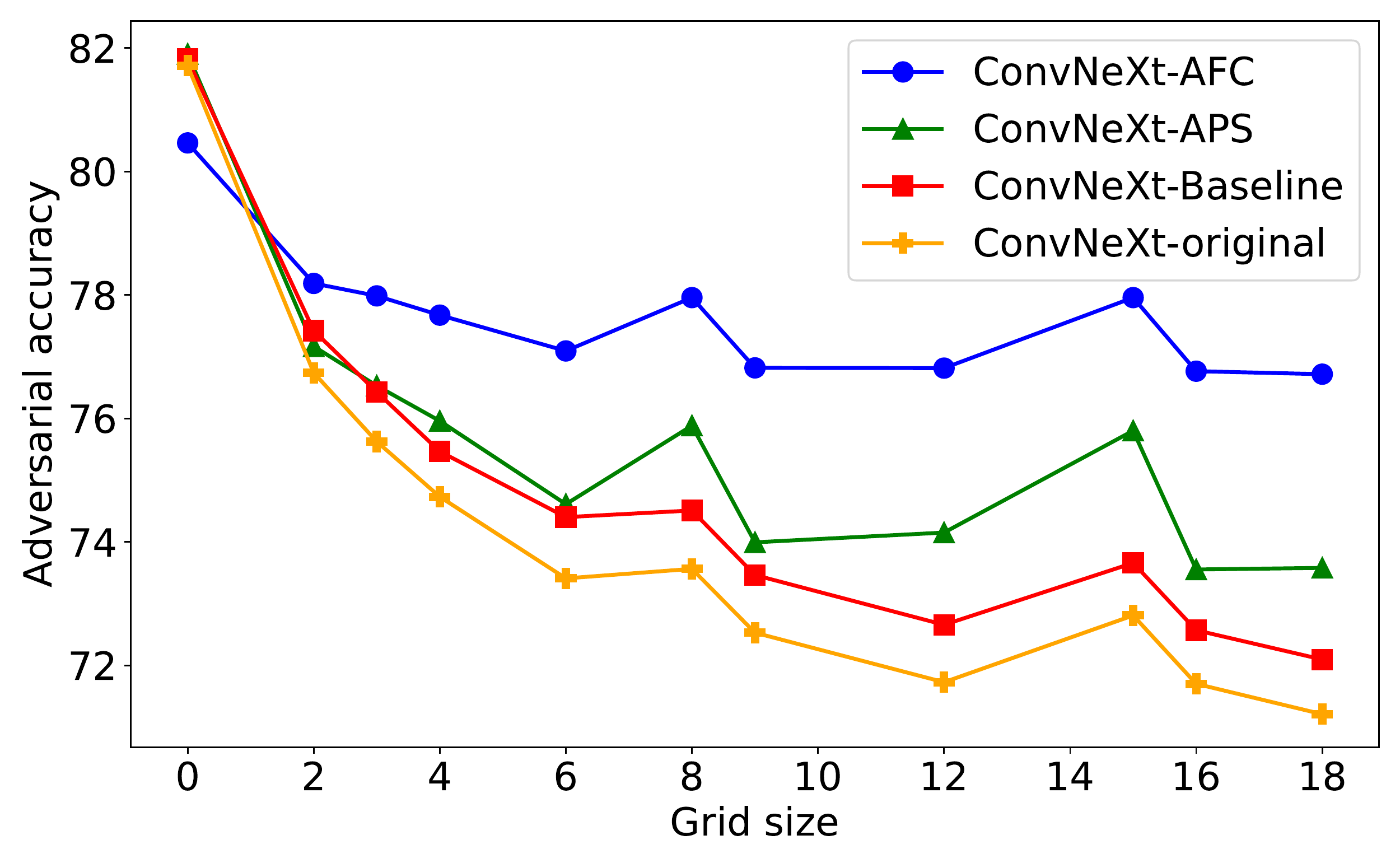}
    \end{subfigure}
    \begin{subfigure}[b]{0.45\linewidth}
    \centering
    \includegraphics[width=\linewidth]{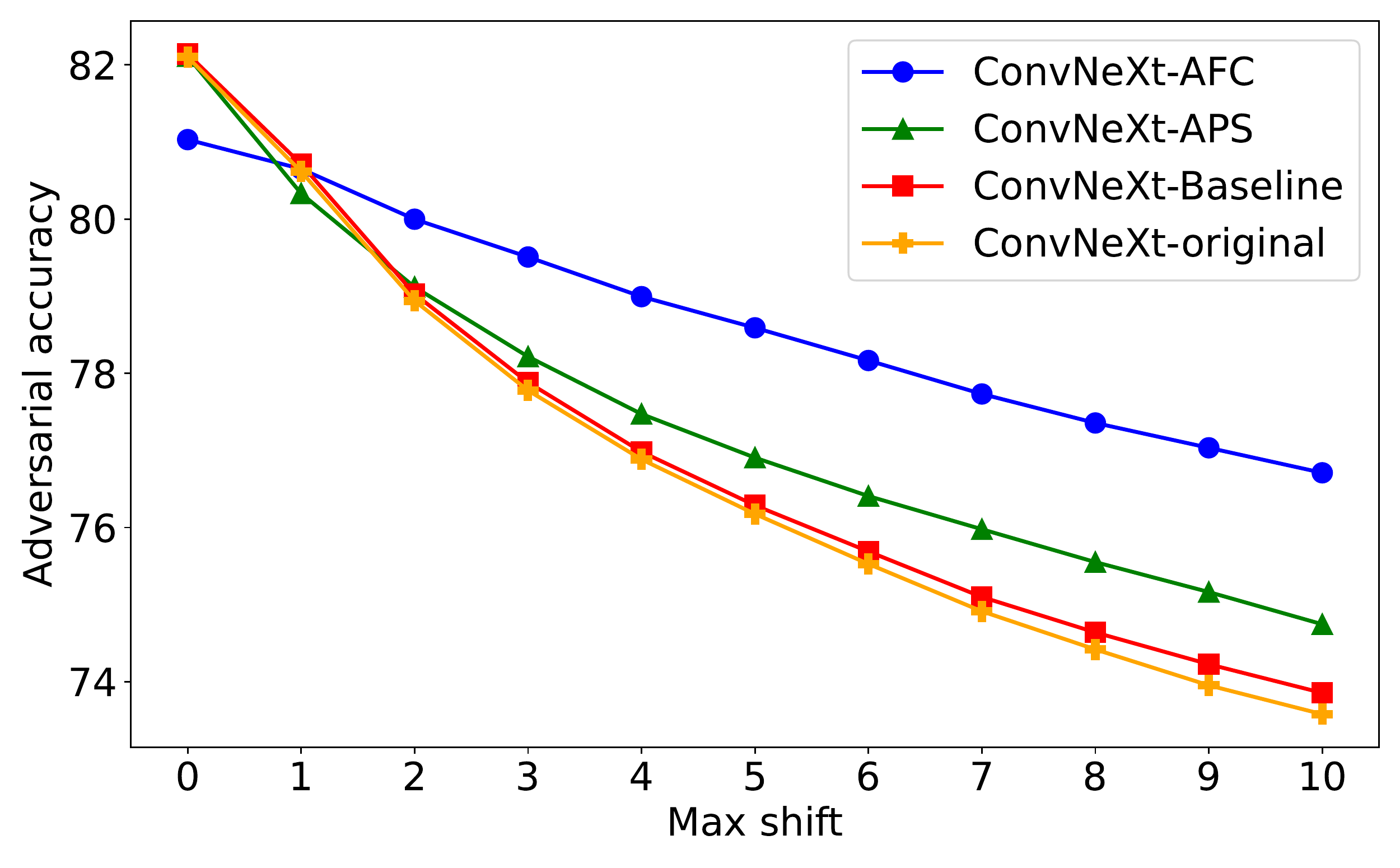}
    \end{subfigure}
\caption{\textbf{Adversarial accuracy for other types of shifts.} \textbf{Left}: Zero-padding, bilinear interpolation results. AFC is the most robust model for all tested grid sizes. 
   \textbf{Right}: Crop-shift results. AFC is the most robust model for $m\geq2$. 
   The accuracy improvement over the baseline and APS models reaches to $2.9\%$ and $2\%$ respectively for the strongest attack in our scope ($m=10$).} 
   \label{fig:crop-shift-bilinear}
   \end{center}
   \vskip -0.2in
\end{figure}

In the experiments above, we used the common ImageNet input: the $224\times224$ center crop of the original $256\times256$ image. 
In contrast, in this experiment, we adversarially translated the cropped area, modeling translating a camera w.r.t.~the scene, as shown in \Cref{fig:circular_crop_shifts}(c).
We measure the adversarial accuracy of translations by up to $m$ integer pixels in each direction (i.e.~grid search at size $\left( 2m+1 \right) \times \left( 2m+1 \right)$). 
The results in \Cref{fig:crop-shift-bilinear} (right) show that our model is more robust to this kind of translation, which is not cyclic, includes data loss, and is even integer-valued.
We additionally evaluate the original ConvNeXt model (zero-pad convolutions) which interestingly has the worst robustness in this setting.

\begin{table}
\caption{\textbf{Test accuracy in models with polynomial activations.} }
  \label{table:polynomial-vit-results}

\small
\centering
\begin{tabular}{llc}
\toprule
Task & Model & Test acc. \\ 
\midrule
ImageNet & ConvNeXt-tiny (GeLU) & 82.1 \\
ImageNet & ConvNeXt-tiny (Poly. deg 2) & 81.98 \\ \\

CIFAR10 & ViT (GeLU) & 97.04 \\
CIFAR10 & ViT (Poly. deg 2) & 97.08 \\
ImageNet & Deit-tiny (GeLU) & 72.29 \\
ImageNet & Deit-tiny (Poly. deg 4) & 71.96 \\

\bottomrule
\end{tabular}


\end{table}

\section{Related work}
Modern Convolutional Neural Networks use downsampling operations such as pooling and strided convolutions to increase the net's receptive field, with lower computation cost than using larger kernels for that matter. 
It was shown that this architectural design breaks the shift-equivariance property of the convolution operation due to the aliasing effect \citep{Azulay2019WhyTransformations}, leading CNNs to be not shift-invariant. 
Even though it was shown this property can be partially learned using appropriate data augmentation \citep{Gunasekar2022GeneralizationAugmentations}, other works tried to architecturally regain shift-invariance. 

Another work \citep{Zhang2019MakingAgain} suggested shift-equivariance could be maintained by reducing aliasing using low-pass filters (LPFs) before downsampling \citep{oppenheim99}. 
Others \citep{Zou2020DelvingConvNets} improved the LPF method by using adaptive content-aware adaptive filters.
These changes were shown to improve convnets robustness to translations, as well as accuracy and generalization, yet another work showed that focusing on circular shifts may induce adversarial attack vulnerabilities \citep{Singla2021ShiftRobustness}.

Instead of tackling the aliasing problem, other works suggested solving shift variance by using adaptive subsampling grids \citep{Chaman2020TrulyNetworks, Rojas-Gomez2022LearnableNetworks, Xu2021GroupSubsampling}. 
This approach was shown ability to produce perfect shift-invariance in image classification tasks. 
Yet, as it does not eliminate the aliasing effects, it does not produce shift-invariance to fractional shifts, and it does not ensure shift-equivariance of the internal representations.

Since it is known that aliasing in discrete signals is caused by non-linearities in addition to subsampling, a few studies suggested methods for alias-free activation functions. 
\citet{Karras2021Alias-FreeNetworks} suggested using upsampling before non-linearities to reduce aliasing in generative models, which cause failure in embedding ``high-frequency features'' such as textures in their outputs. 
The idea of using polynomial non-linearities to battle aliasing has been mentioned previously \citep{oppenheim99, EmmyWei2022Aliasing-FreeFunctions}. \citet{Franzen2021GeneralCNNs} have recently shown this methodology can be used to improve rotation-equivariance. 
However, it has never been applied in a complete alias-free setting, nor in modern-scale deep networks. 
Other smooth activation functions have been suggested as well \citep{Hossain2021Anti-aliasingFunction, Vasconcelos2020AnNetworks}, yet they do not completely eliminate aliasing.

It is worth mentioning that other equivariance properties have been studied as well, such as rotation, reflection and group equivariance \citep{Delchevalerie2021AchievingNetworks, Bronstein2016GeometricData, Xu2021GroupSubsampling, Ning2022Scale-AwareEquivariance, Manfredi2020ShiftDetection, Romero2020AttentiveNetworks, Yeh2022EquivarianceParameter-Sharing, Weiler2019GeneralCNNs1,
Delchevalerie2021AchievingNetworks}. This work is focused on the specific property of shift-invariance in CNNs for image classification.
\section{Discussion}
In this paper we proposed the Alias-Free Convnet, which for the first time, is guaranteed to eliminate any aliasing effects in the model, to ensure the output is invariant to any input shifts (even sub-pixel ones), and to ensure the internal representations are equivariant to any shifts (even sub-pixel ones). 
We demonstrate this numerically and show this leads to (certified) high performance under adversarial shift-based attacks --- in contrast to existing models which degrade in performance. 
However, this comes at a cost, such as a 1.08\% reduction in standard test accuracy (as methods that increase robustness often reduce accuracy). 

\subsection{Accuracy}
Although our model has improved robustness in comparison to the baseline model and other shift-invariance methods, it has a lower accuracy on the original test set. 
This result makes sense since robustness often comes at the cost of accuracy. 
Specifically, the property of perfect shift-invariance is architecturally forced in our model, in contrast to other CNNs where it can be violated. This may seem as a reduction of the hypothesis set.
However, in \Cref{table:aal-modifcations-accuracy} we observe that the highest drop in accuracy does not occur at the last modification, where the model becomes shift-invariant, but rather as a result of modifying the Normalization Layer to be alias-free.
Although the proposed alias-free normalization layer was designed to remain similar to the original model LayerNorm, other alias-free normalization methods may exist and lead to a higher accuracy than ours, without hurting robustness. 
Another drop in accuracy occurs as a result of replacing GeLU in polynomial activation, which surprisingly does not happen in the regular setting (models without cyclic convolutions); a polynomial activation in a cyclic setting leads to a reduction of 0.4\% in accuracy (\cref{table:aal-modifcations-accuracy}) while in a non-cyclic setting it leads to a reduction of 0.1\% only (\cref{table:polynomial-vit-results}). 
It implies that additional hyper-parameter tuning might help in the recovery of this accuracy drop.
In addition, we note again that the modifications in our model effectively remove non-linearities from the baseline model, and that ConvNeXt-AFC is practically a polynomial of the input with a degree that is an exponent of the convnet's depth.  
Thus, using a wider or deeper convnet might help in closing the accuracy gap.


\subsection{Runtime performance}
Although the AFC model has only a small amount of additional parameters in the polynomial activation function, it has a higher computation cost than the baseline (see \cref{table:baenchmark-results}). The main reason for that is that while the activation function in the baseline model is a single pointwise operation, our activation requires upsampling and downsampling which are rather expensive. This issue has been addressed in a previous study \citep{Karras2021Alias-FreeNetworks}, where a similar scheme for partially alias-free activation has been used. They combined all the required operations to a single CUDA kernel which (according to them) led to a speed-up of at least x20 over native PyTorch implementation, and in total to  x10 speed-up in training time.
Our model training time is only 5 times higher than the baseline training time, therefore it is reasonable to assume such efficient implementation may significantly reduce the training time gap.

\begin{table}[h]
\caption{\textbf{Training and evaluation performance.} Train time was measured on Nvidia A6000 x 8 using the maximal possible batch-size per model due to memory constraints. Evaluation time was measured on a single A6000 with batch-size 256.}
  \label{table:baenchmark-results}

  \centering
\resizebox{1.0\linewidth}{!}{
    \begin{tabular}{lll}
    \toprule

Model & Train time [hours] & Eval time [ms per sample] \\ 
\midrule

ConvNeXt-Baseline \cite{Liu2022A2020s}           & 84        &  1.39  \\
ConvNeXt-APS   \cite{Chaman2020TrulyNetworks}    & 93     &   1.56  \\
ConvNeXt-AFC  (ours)                          & 418    &  9.16 \\
\bottomrule
  \end{tabular}
  }
\end{table}

\subsection{Image translations}
\subsubsection{Circular translations}
\begin{figure}[ht]
    \centering
    \begin{subfigure}[b]{0.32\linewidth}
    \centering
    \includegraphics[width=\linewidth]{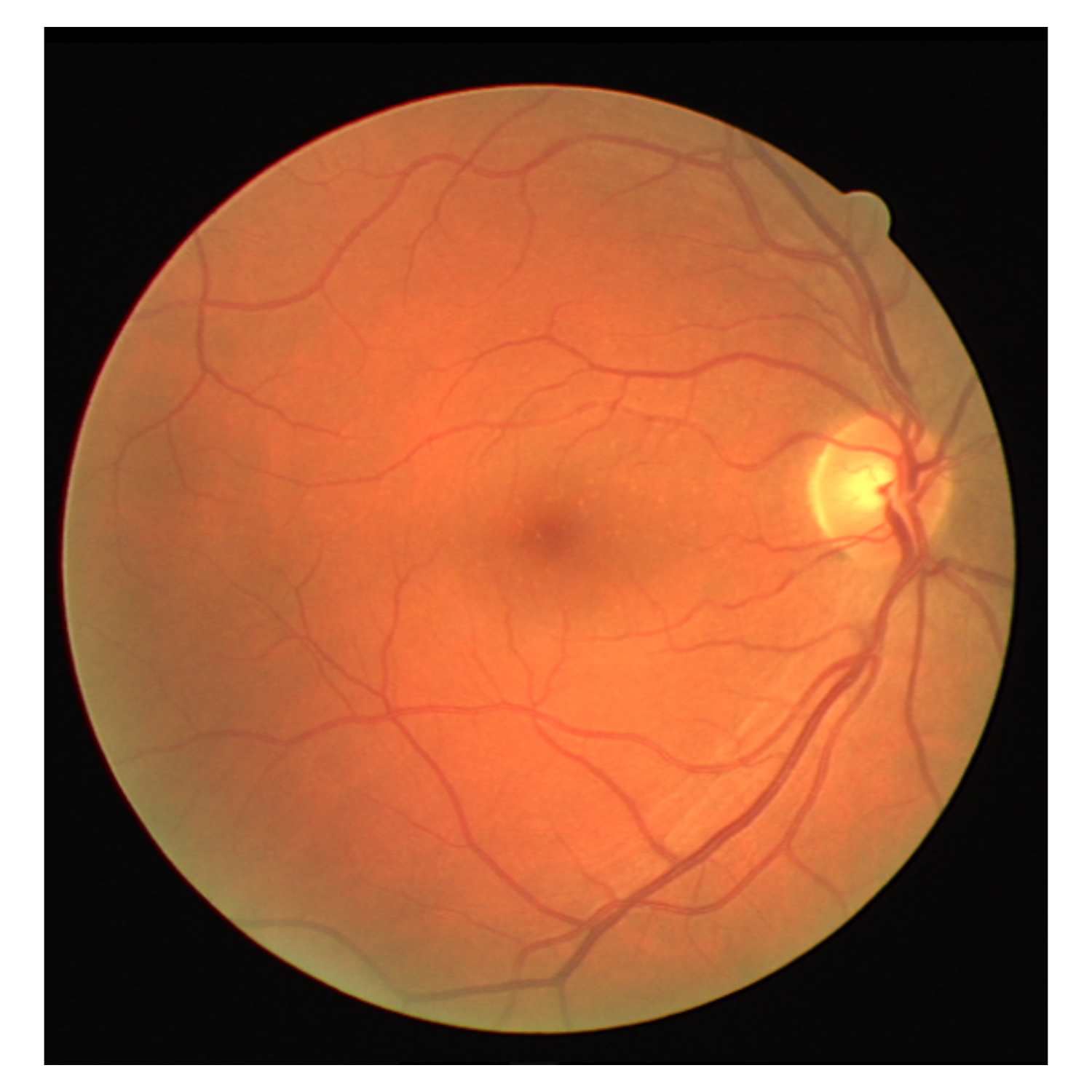}
    \caption{Original image}
    \end{subfigure}
    \begin{subfigure}[b]{0.32\linewidth}
    \centering
    \includegraphics[width=\linewidth]{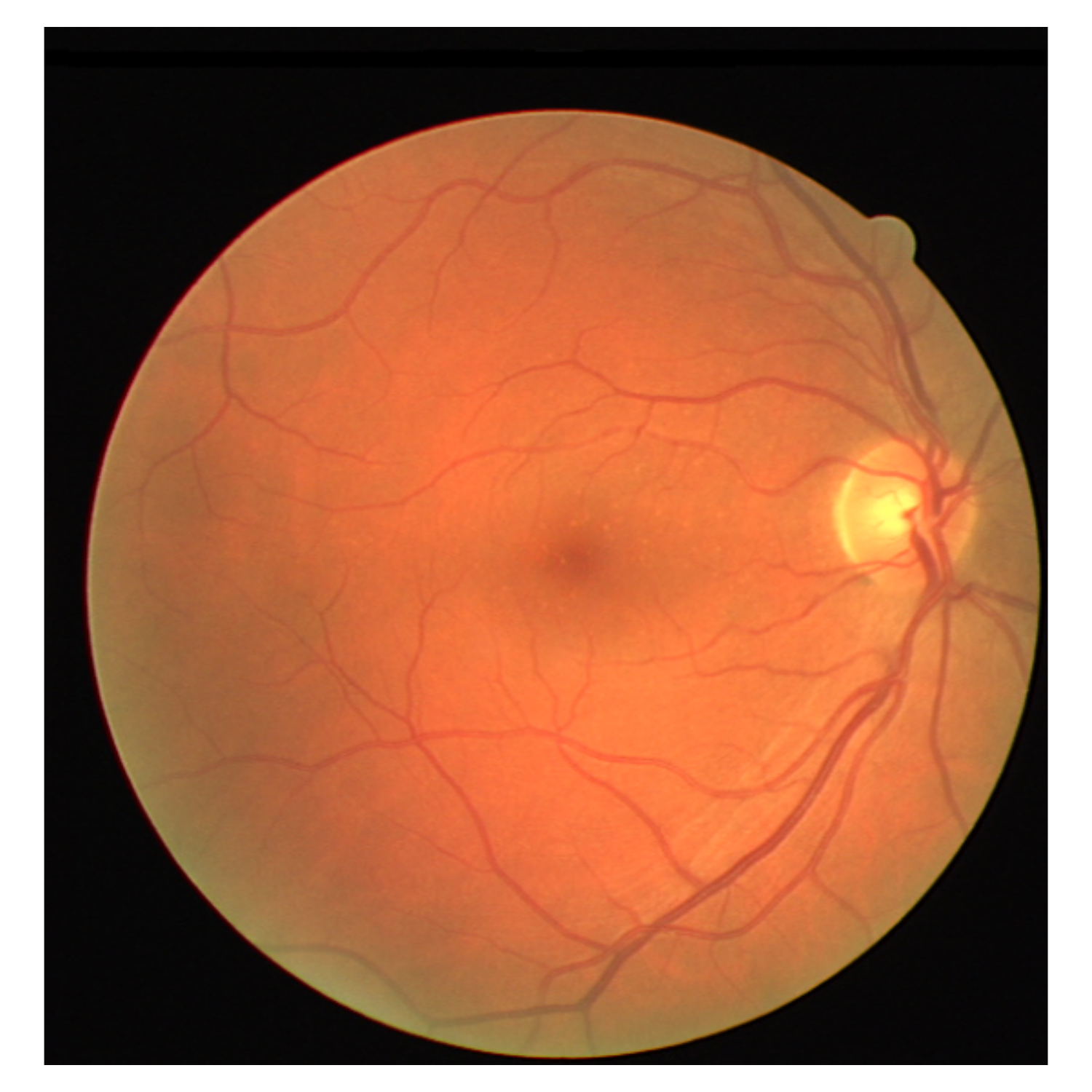}
    \caption{Circular shifted image}
    \end{subfigure}
\caption{
 \textbf{ Circular translation of a retinal image.} (a) Original Image \citep{Staal2004Ridge-BasedRetina}. (b) Circular translated image. The retinal image has a uniform background, hence circular shift is equivalent to a translation of the object within the image (i.e.~``crop-shift'').
  }
\label{fig:medical_shifts}
\end{figure}

\begin{figure}[ht]
    \centering
    \begin{subfigure}[b]{0.8\linewidth}
    \centering
    \includegraphics[width=\linewidth]{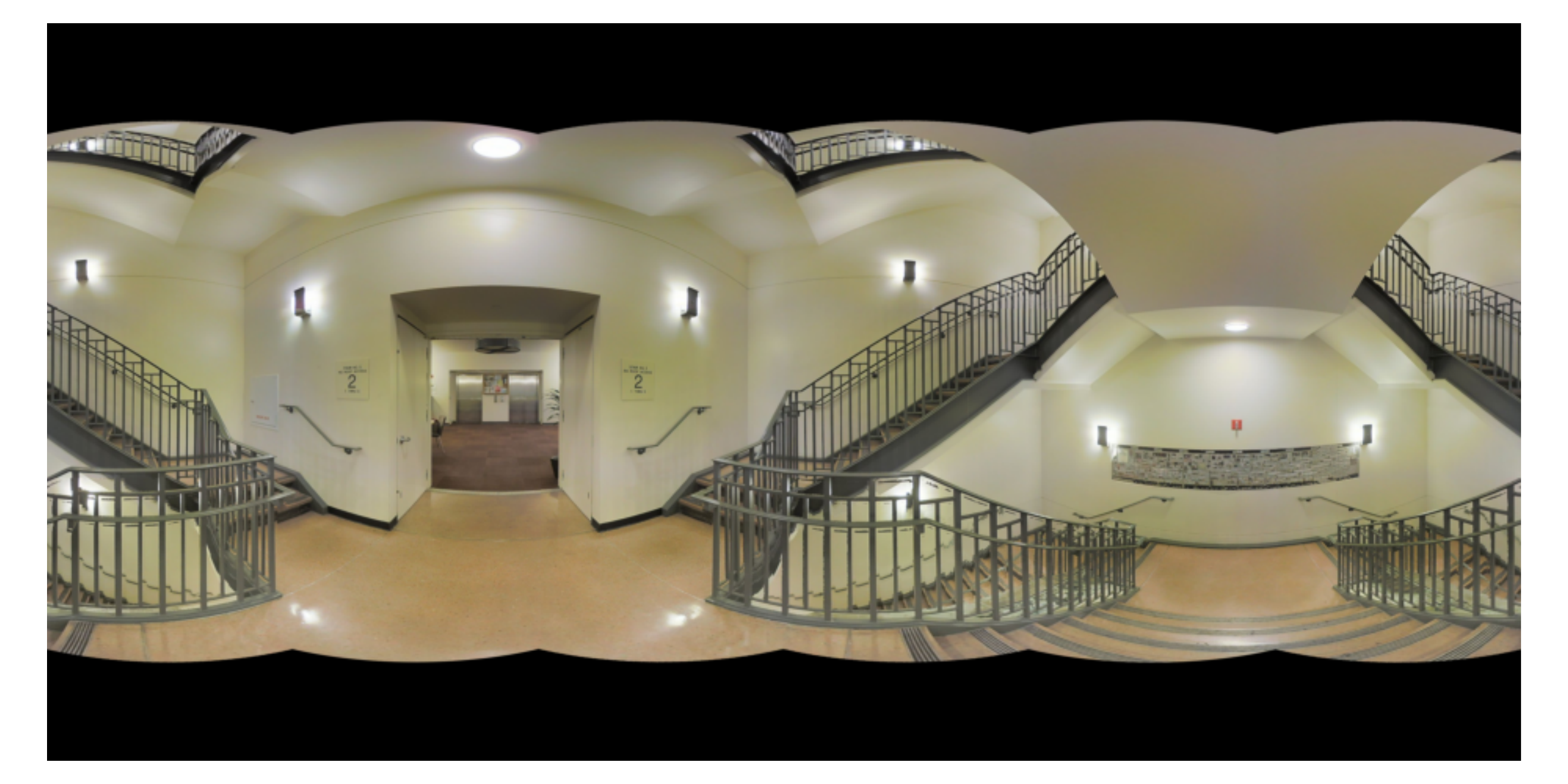}
    \caption{Original image}
    \end{subfigure}
    \begin{subfigure}[b]{0.8\linewidth}
    \centering
    \includegraphics[width=\linewidth]{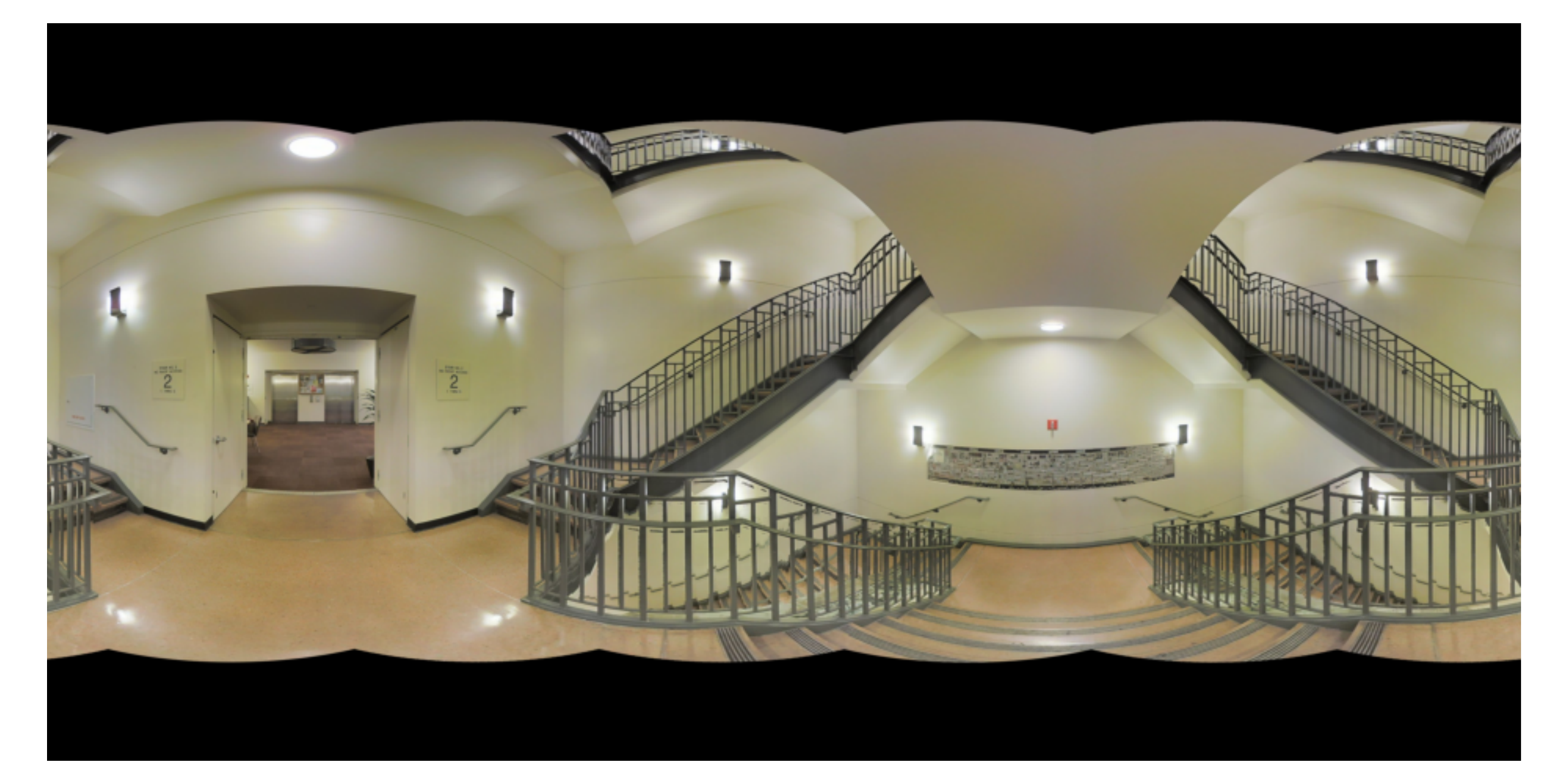}
    \caption{Circular shifted image}
    \end{subfigure}
\caption{ \textbf{Circular translation of a panoramic image.} (a) Original Image \citep{Armeni2017JointUnderstanding}. (b) Circular translated image, representing a translation of the camera with respect to the scene, without any edge artifacts.
    }
\label{fig:panoram_shifts}
\end{figure}

The guaranteed robustness in the AFC model is limited to circular shifts, similarly to previous work \citep{Chaman2020TrulyNetworks}. 
Applying this kind of translation on a finite image causes edge artifacts and creates an unrealistic image (e.g., see \cref{fig:circular_crop_shifts}).
Although our model has improved robustness even in translations of the frame with respect to the scene (\cref{fig:crop-shift-bilinear}), which may seem as a more practical setting, information-loss makes guaranteed robustness impossible --- for example, consider an image in which a translation cause the classified object to get out of the frame. 
Although the certified robustness may seem not applicable, circular shifts can actually be practically relevant. 
For example, shifting an object over a constant (i.e., uniform) background will seem identical to a circular translation of the entire frame. 
This setting may be relevant in face recognition tasks and medical imaging (see \cref{fig:medical_shifts}). 
In addition, horizontal circular shifts are relevant for panoramic (360$^{\circ}$) cameras, e.g.~in autonomous cars (see \cref{fig:panoram_shifts}).

\subsubsection{Interpolation kernel}
In \Cref{sec:methods} we proved that AFC is robust even to sub-pixel shifts. 
Our robustness guarantees assume the digital image processed by the network corresponds to point-wise samples of a continuous-space image that had been convolved with a perfect anti-aliasing filter prior to sampling (though, empirically, our method performs well with other types of interpolation kernels, e.g. \cref{fig:crop-shift-bilinear}). 
Although this may seem like a serious limitation, it is in fact the standard setting in any imaging system. 
Indeed, in any optical imaging system, the image impinging on the detector corresponds to the continuous scene convolved with a low-pass filter. 
This low-pass filter completely zeros out all frequencies above a cutoff frequency that is inversely proportional to the diameter of the aperture 
\citep{Goodman1969IntroductionOptics}. Thus, while in many domains perfect low-pass filtering is challenging to achieve, in Optics, this diffraction limit is in fact impossible to avoid. 
Cameras are designed such that the cutoff frequency of the aperture corresponds to the Nyquist frequency of the CCD array. 
In systems where the aperture can be modified by the user, the CCD array is typically adjusted to the minimal aperture width and an additional anti-aliasing filter element is inserted in front of the CCD \citep{Schoberl2010DimensioningCameras}, so that the Nyquist condition is still met for any chosen aperture diameter within the allowed range. It should be noted, however, that the digital image captured by the sensor typically undergoes a series of nonlinear operations within the image signal processor (ISP) of the camera.
This implies that shift equivariance may be lost already at the camera level, before the image reaches our convnet. Addressing these effects is beyond the scope of our work. 


\subsection{Polynomial activation functions}
Polynomial activation functions are relatively cheap to compute, which might motivate their use in the future, regardless of their advantage in the context of aliasing-free convnets. 
Despite this, the high curvature of the polynomial activation seems a barrier to its usage in a wider variety of architectures. 
To achieve good performance on CIFAR, \citet{Gottemukkula2019POLYNOMIALFUNCTIONS} implemented in ResNet \citep{He2015DeepRecognition} a stable version of polynomial activation function, by scaling the pre-activation by the $L_1$ of the layer, which bounds the maximal input element (note that this scaling technique is not aliasing-free and thus was not useful for our purposes). 
We observe this scaling is unnecessary in architectures with sparser usage of activation functions, such as ConvNeXt and ViT; in \Cref{table:polynomial-vit-results}, we show preliminary results suggesting a polynomial activation may be a reasonable substitute for the widely spread GeLU in Transformer-based architectures as well as convnets. 

\subsection{Future work}
This study shows how an aliasing-free convnet can be used to build an image classifier with certified shift robustness. 
Yet, the applications of such convnet are not limited to that purpose; our method can be applied in other domains in which aliasing has been shown to be damaging, such as generative models \citep{Karras2021Alias-FreeNetworks}. 
Furthermore, our method guarantees shift-equivariant internal representation, a stronger property than shift-invariance. 
Future work may examine the importance of this property in other tasks. For example, our method can be naturally expanded to construct a shift-equivariant convnets for segmentation. 

\subsubsection*{Acknowledgments}
 The research of DS was funded by the European Union (ERC, A-B-C-Deep, 101039436). Views and opinions expressed are however those of the author only and do not necessarily reflect those of the European Union or the European Research Council Executive Agency (ERCEA). Neither the European Union nor the granting authority can be held responsible for them. DS also acknowledges the support of Schmidt Career Advancement Chair in AI. TM was supported by grant 2318/22 from the Israel Science Foundation and by the Ollendorff Center of the Viterbi Faculty of Electrical and Computer Engineering at the Technion.

\bibliography{references, references_manual}  

\begin{thebibliography}{34}
\providecommand{\natexlab}[1]{#1}
\providecommand{\url}[1]{\texttt{#1}}
\expandafter\ifx\csname urlstyle\endcsname\relax
  \providecommand{\doi}[1]{doi: #1}\else
  \providecommand{\doi}{doi: \begingroup \urlstyle{rm}\Url}\fi

\bibitem[Armeni et~al.(2017)Armeni, Sax, Zamir, and
  Savarese]{Armeni2017JointUnderstanding}
I.~Armeni, S.~Sax, A.~R. Zamir, and S.~Savarese.
\newblock {Joint 2D-3D-Semantic Data for Indoor Scene Understanding}.
\newblock 2 2017.

\bibitem[Azulay and Weiss(2019)]{Azulay2019WhyTransformations}
A.~Azulay and Y.~Weiss.
\newblock {Why do deep convolutional networks generalize so poorly to small
  image transformations?}
\newblock \emph{Journal of Machine Learning Research}, 20:\penalty0 1--25,
  2019.
\newblock URL \url{https://youtu.be/MpUdRacvkWk}.

\bibitem[Bronstein et~al.(2016)Bronstein, Bruna, Lecun, Szlam, and
  Vandergheynst]{Bronstein2016GeometricData}
M.~M. Bronstein, J.~Bruna, Y.~Lecun, A.~Szlam, and P.~Vandergheynst.
\newblock {Geometric deep learning: going beyond Euclidean data}.
\newblock \emph{IEEE Signal Processing Magazine}, 34\penalty0 (4):\penalty0
  18--42, 11 2016.
\newblock ISSN 10535888.
\newblock \doi{10.1109/msp.2017.2693418}.
\newblock URL \url{https://arxiv.org/abs/1611.08097v2}.

\bibitem[Chaman and Dokmani{\'{c}}(2020)]{Chaman2020TrulyNetworks}
A.~Chaman and I.~Dokmani{\'{c}}.
\newblock {Truly shift-invariant convolutional neural networks}.
\newblock \emph{Proceedings of the IEEE Computer Society Conference on Computer
  Vision and Pattern Recognition}, pages 3772--3782, 11 2020.
\newblock ISSN 10636919.
\newblock \doi{10.48550/arxiv.2011.14214}.
\newblock URL \url{https://arxiv.org/abs/2011.14214v4}.

\bibitem[Delchevalerie et~al.(2021)Delchevalerie, Bibal, Fr{\'{e}}nay, and
  Mayer]{Delchevalerie2021AchievingNetworks}
V.~Delchevalerie, A.~Bibal, B.~Fr{\'{e}}nay, and A.~Mayer.
\newblock {Achieving Rotational Invariance with Bessel-Convolutional Neural
  Networks}.
\newblock \emph{Advances in Neural Information Processing Systems},
  34:\penalty0 28772--28783, 12 2021.

\bibitem[Deng et~al.(2009)Deng, Dong, Socher, Li, {Kai Li}, and {Li
  Fei-Fei}]{Deng2009ImageNet:Database}
J.~Deng, W.~Dong, R.~Socher, L.-J. Li, {Kai Li}, and {Li Fei-Fei}.
\newblock {ImageNet: A large-scale hierarchical image database}.
\newblock In \emph{2009 IEEE Conference on Computer Vision and Pattern
  Recognition}, pages 248--255. IEEE, 6 2009.
\newblock ISBN 978-1-4244-3992-8.
\newblock \doi{10.1109/CVPR.2009.5206848}.

\bibitem[Emmy~Wei(2022)]{EmmyWei2022Aliasing-FreeFunctions}
S.~Emmy~Wei.
\newblock {Aliasing-Free Nonlinear Signal Processing Using Implicitly Defined
  Functions}.
\newblock \emph{IEEE Access}, 10:\penalty0 76281--76295, 2022.
\newblock ISSN 21693536.
\newblock \doi{10.1109/ACCESS.2022.3192387}.

\bibitem[Engstrom et~al.(2017)Engstrom, Tran, Tsipras, Schmidt, and
  Madry]{Engstrom2017ExploringRobustness}
L.~Engstrom, B.~Tran, D.~Tsipras, L.~Schmidt, and A.~Madry.
\newblock {Exploring the Landscape of Spatial Robustness}.
\newblock \emph{36th International Conference on Machine Learning, ICML 2019},
  2019-June:\penalty0 3218--3238, 12 2017.
\newblock \doi{10.48550/arxiv.1712.02779}.
\newblock URL \url{https://arxiv.org/abs/1712.02779v4}.

\bibitem[Franzen and Wand(2021)]{Franzen2021GeneralCNNs}
D.~Franzen and M.~Wand.
\newblock {General Nonlinearities in SO(2)-Equivariant CNNs}.
\newblock \emph{Advances in Neural Information Processing Systems},
  34:\penalty0 9086--9098, 12 2021.

\bibitem[Goodman and Cox(1969)]{Goodman1969IntroductionOptics}
J.~W. Goodman and M.~E. Cox.
\newblock {Introduction to Fourier Optics}.
\newblock \emph{Physics Today}, 22\penalty0 (4):\penalty0 97--101, 4 1969.
\newblock ISSN 0031-9228.
\newblock \doi{10.1063/1.3035549}.

\bibitem[Gottemukkula(2019)]{Gottemukkula2019POLYNOMIALFUNCTIONS}
V.~Gottemukkula.
\newblock {POLYNOMIAL ACTIVATION FUNCTIONS}.
\newblock Technical report, 2019.

\bibitem[Gunasekar(2022)]{Gunasekar2022GeneralizationAugmentations}
S.~Gunasekar.
\newblock {Generalization to translation shifts: a study in architectures and
  augmentations}, 7 2022.
\newblock URL \url{https://arxiv.org/abs/2207.02349v1}.

\bibitem[He et~al.(2015)He, Zhang, Ren, and Sun]{He2015DeepRecognition}
K.~He, X.~Zhang, S.~Ren, and J.~Sun.
\newblock {Deep Residual Learning for Image Recognition}.
\newblock \emph{Proceedings of the IEEE Computer Society Conference on Computer
  Vision and Pattern Recognition}, 2016-December:\penalty0 770--778, 12 2015.
\newblock ISSN 10636919.
\newblock \doi{10.48550/arxiv.1512.03385}.
\newblock URL \url{https://arxiv.org/abs/1512.03385v1}.

\bibitem[Hendrycks and
  Dietterich(2019)]{Hendrycks2019BenchmarkingPerturbations}
D.~Hendrycks and T.~Dietterich.
\newblock {Benchmarking Neural Network Robustness to Common Corruptions and
  Perturbations}.
\newblock \emph{7th International Conference on Learning Representations, ICLR
  2019}, 3 2019.
\newblock \doi{10.48550/arxiv.1903.12261}.
\newblock URL \url{https://arxiv.org/abs/1903.12261v1}.

\bibitem[Hornik et~al.(1989)Hornik, Stinchcombe, and
  White]{Hornik1989MultilayerApproximators}
K.~Hornik, M.~Stinchcombe, and H.~White.
\newblock {Multilayer feedforward networks are universal approximators}.
\newblock \emph{Neural Networks}, 2\penalty0 (5):\penalty0 359--366, 1 1989.
\newblock ISSN 08936080.
\newblock \doi{10.1016/0893-6080(89)90020-8}.

\bibitem[Hossain et~al.(2021)Hossain, Teng, Sohel, and
  Lu]{Hossain2021Anti-aliasingFunction}
M.~T. Hossain, S.~W. Teng, F.~Sohel, and G.~Lu.
\newblock {Anti-aliasing Deep Image Classifiers using Novel Depth Adaptive
  Blurring and Activation Function}, 10 2021.
\newblock URL \url{https://arxiv.org/abs/2110.00899v1}.

\bibitem[Karras et~al.(2021)Karras, Aittala, Laine, H{\"{a}}rk{\"{o}}nen,
  Hellsten, Lehtinen, and Aila]{Karras2021Alias-FreeNetworks}
T.~Karras, M.~Aittala, S.~Laine, E.~H{\"{a}}rk{\"{o}}nen, J.~Hellsten,
  J.~Lehtinen, and T.~Aila.
\newblock {Alias-Free Generative Adversarial Networks}, 6 2021.

\bibitem[Kidger and Lyons(2020)]{Kidger2020UniversalNetworks}
P.~Kidger and T.~Lyons.
\newblock {Universal Approximation with Deep Narrow Networks}.
\newblock \emph{Proceedings of Machine Learning Research}, TBD:\penalty0 1--22,
  2020.

\bibitem[Liu et~al.(2022)Liu, Mao, Wu, Feichtenhofer, Darrell, Xie, and
  Research]{Liu2022A2020s}
Z.~Liu, H.~Mao, C.-Y. Wu, C.~Feichtenhofer, T.~Darrell, S.~Xie, and F.~A.
  Research.
\newblock {A ConvNet for the 2020s}.
\newblock In \emph{Proceedings of the IEEE/CVF Conference on Computer Vision
  and Pattern Recognition}, pages 11976--11986, 2022.
\newblock URL \url{https://github.com/facebookresearch/ConvNeXt}.

\bibitem[Manfredi and Wang(2020)]{Manfredi2020ShiftDetection}
M.~Manfredi and Y.~Wang.
\newblock {Shift Equivariance in Object Detection}.
\newblock \emph{Lecture Notes in Computer Science (including subseries Lecture
  Notes in Artificial Intelligence and Lecture Notes in Bioinformatics)}, 12540
  LNCS:\penalty0 32--45, 8 2020.
\newblock ISSN 16113349.
\newblock \doi{10.48550/arxiv.2008.05787}.
\newblock URL \url{https://arxiv.org/abs/2008.05787v1}.

\bibitem[Ning et~al.(2022)Ning, Tang, Zhong, Wu, Zhang, and
  Zhang]{Ning2022Scale-AwareEquivariance}
M.~Ning, J.~Tang, H.~Zhong, H.~Wu, P.~Zhang, and Z.~Zhang.
\newblock {Scale-Aware Network with Scale Equivariance}.
\newblock \emph{Photonics 2022, Vol. 9, Page 142}, 9\penalty0 (3):\penalty0
  142, 2 2022.
\newblock ISSN 2304-6732.
\newblock \doi{10.3390/PHOTONICS9030142}.
\newblock URL \url{https://www.mdpi.com/2304-6732/9/3/142/htm
  https://www.mdpi.com/2304-6732/9/3/142}.

\bibitem[Oppenheim et~al.(1999)Oppenheim, Schafer, and Buck]{oppenheim99}
A.~V. Oppenheim, R.~W. Schafer, and J.~R. Buck.
\newblock \emph{Discrete-Time Signal Processing}.
\newblock Prentice-hall Englewood Cliffs, second edition, 1999.

\bibitem[Paszke et~al.(2019)Paszke, Gross, Massa, Lerer, Bradbury, Chanan,
  Killeen, Lin, Gimelshein, Antiga, Desmaison, Kopf, Yang, DeVito, Raison,
  Tejani, Chilamkurthy, Steiner, Fang, Bai, and Chintala]{NEURIPS2019_9015}
A.~Paszke, S.~Gross, F.~Massa, A.~Lerer, J.~Bradbury, G.~Chanan, T.~Killeen,
  Z.~Lin, N.~Gimelshein, L.~Antiga, A.~Desmaison, A.~Kopf, E.~Yang, Z.~DeVito,
  M.~Raison, A.~Tejani, S.~Chilamkurthy, B.~Steiner, L.~Fang, J.~Bai, and
  S.~Chintala.
\newblock Pytorch: An imperative style, high-performance deep learning library.
\newblock In \emph{Advances in Neural Information Processing Systems 32}, pages
  8024--8035. Curran Associates, Inc., 2019.
\newblock URL
  \url{http://papers.neurips.cc/paper/9015-pytorch-an-imperative-style-high-performance-deep-learning-library.pdf}.

\bibitem[Rojas-Gomez et~al.(2022)Rojas-Gomez, Lim, Schwing, Do, and
  Yeh]{Rojas-Gomez2022LearnableNetworks}
R.~A. Rojas-Gomez, T.-Y. Lim, A.~G. Schwing, M.~N. Do, and R.~A. Yeh.
\newblock {Learnable Polyphase Sampling for Shift Invariant and Equivariant
  Convolutional Networks}.
\newblock 10 2022.
\newblock \doi{10.48550/arxiv.2210.08001}.
\newblock URL \url{https://arxiv.org/abs/2210.08001v1}.

\bibitem[Romero et~al.(2020)Romero, Bekkers, Tomczak, and
  Hoogendoorn]{Romero2020AttentiveNetworks}
D.~W. Romero, E.~J. Bekkers, J.~M. Tomczak, and M.~Hoogendoorn.
\newblock {Attentive Group Equivariant Convolutional Networks}.
\newblock 2 2020.
\newblock \doi{10.48550/arxiv.2002.03830}.
\newblock URL \url{https://arxiv.org/abs/2002.03830v3}.

\bibitem[Schoberl et~al.(2010)Schoberl, Schnurrer, Oberdorster, Fossel, and
  Kaup]{Schoberl2010DimensioningCameras}
M.~Schoberl, W.~Schnurrer, A.~Oberdorster, S.~Fossel, and A.~Kaup.
\newblock {Dimensioning of optical birefringent anti-alias filters for digital
  cameras}.
\newblock In \emph{2010 IEEE International Conference on Image Processing},
  pages 4305--4308. IEEE, 9 2010.
\newblock ISBN 978-1-4244-7992-4.
\newblock \doi{10.1109/ICIP.2010.5651784}.

\bibitem[Singla et~al.(2021)Singla, Ge, Basri, and
  Jacobs]{Singla2021ShiftRobustness}
V.~Singla, S.~Ge, R.~Basri, and D.~Jacobs.
\newblock {Shift Invariance Can Reduce Adversarial Robustness}, 3 2021.
\newblock ISSN 10495258.
\newblock URL \url{https://arxiv.org/abs/2103.02695v3}.

\bibitem[Staal et~al.(2004)Staal, Abramoff, Niemeijer, Viergever, and van
  Ginneken]{Staal2004Ridge-BasedRetina}
J.~Staal, M.~Abramoff, M.~Niemeijer, M.~Viergever, and B.~van Ginneken.
\newblock {Ridge-Based Vessel Segmentation in Color Images of the Retina}.
\newblock \emph{IEEE Transactions on Medical Imaging}, 23\penalty0
  (4):\penalty0 501--509, 4 2004.
\newblock ISSN 0278-0062.
\newblock \doi{10.1109/TMI.2004.825627}.

\bibitem[Vasconcelos et~al.(2020)Vasconcelos, Larochelle, Dumoulin, Roux, and
  Goroshin]{Vasconcelos2020AnNetworks}
C.~Vasconcelos, H.~Larochelle, V.~Dumoulin, N.~L. Roux, and R.~Goroshin.
\newblock {An Effective Anti-Aliasing Approach for Residual Networks}.
\newblock 11 2020.
\newblock \doi{10.48550/arxiv.2011.10675}.
\newblock URL \url{https://arxiv.org/abs/2011.10675v1}.

\bibitem[Weiler and Cesa(2019)]{Weiler2019GeneralCNNs1}
M.~Weiler and G.~Cesa.
\newblock {General $E(2)$-Equivariant Steerable CNNs}.
\newblock \emph{Advances in Neural Information Processing Systems}, 32, 11
  2019.
\newblock ISSN 10495258.
\newblock \doi{10.48550/arxiv.1911.08251}.
\newblock URL \url{https://arxiv.org/abs/1911.08251v2}.

\bibitem[Xu et~al.(2021)Xu, Kim, Rainforth, and Teh]{Xu2021GroupSubsampling}
J.~Xu, H.~Kim, T.~Rainforth, and Y.~W. Teh.
\newblock {Group Equivariant Subsampling}.
\newblock \emph{Advances in Neural Information Processing Systems}, 8:\penalty0
  5934--5946, 6 2021.
\newblock ISSN 10495258.
\newblock \doi{10.48550/arxiv.2106.05886}.
\newblock URL \url{https://arxiv.org/abs/2106.05886v1}.

\bibitem[Yeh et~al.(2022)Yeh, Hu, Hasegawa-Johnson, and
  Schwing]{Yeh2022EquivarianceParameter-Sharing}
R.~A. Yeh, Y.-T. Hu, M.~Hasegawa-Johnson, and A.~G. Schwing.
\newblock {Equivariance Discovery by Learned Parameter-Sharing}.
\newblock 2022.

\bibitem[Zhang(2019)]{Zhang2019MakingAgain}
R.~Zhang.
\newblock {Making convolutional networks shift-invariant again}.
\newblock \emph{36th International Conference on Machine Learning, ICML 2019},
  2019-June:\penalty0 12712--12722, 2019.
\newblock URL \url{https://richzhang.github.io/antialiased-cnns/.}

\bibitem[Zou et~al.(2020)Zou, Xiao, Yu, and Lee]{Zou2020DelvingConvNets}
X.~Zou, F.~Xiao, Z.~Yu, and Y.~J. Lee.
\newblock {Delving Deeper into Anti-aliasing in ConvNets}.
\newblock \emph{International Journal of Computer Vision 2022}, pages 1--15, 8
  2020.
\newblock ISSN 15731405.
\newblock \doi{10.1007/S11263-022-01672-Y/FIGURES/11}.
\newblock URL \url{http://arxiv.org/abs/2008.09604}.

\end{thebibliography}


\newpage
\appendix

\section{Polynomial activation function} \label{sec:poly-append}
\subsection{Coefficients initialization}
\begin{figure}[ht]
    \centering
    \begin{subfigure}[b]{0.45\linewidth}
    \centering
    \includegraphics[width=\linewidth]{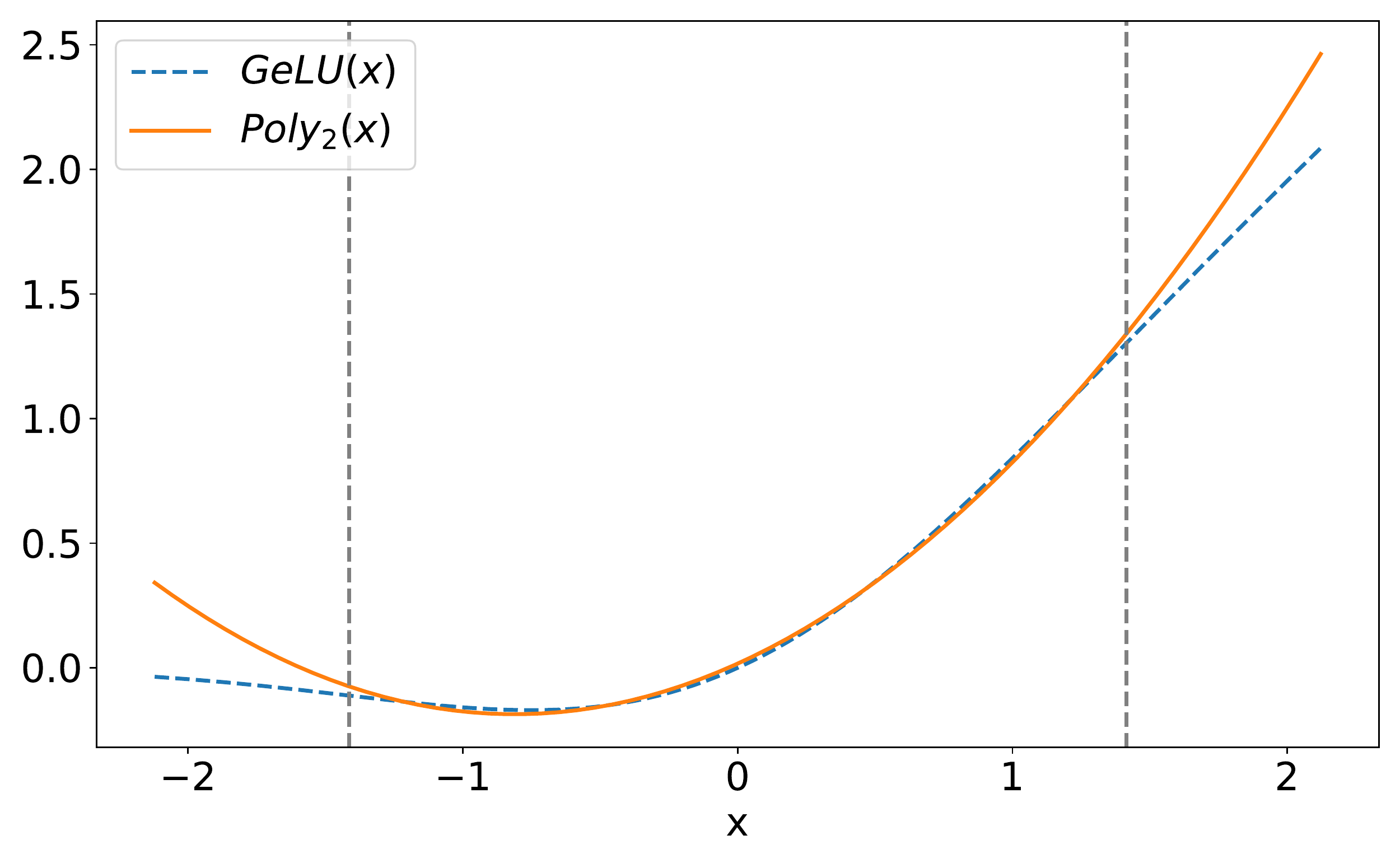}
    \end{subfigure}
    \begin{subfigure}[b]{0.45\linewidth}
    \centering
    \includegraphics[width=\linewidth]{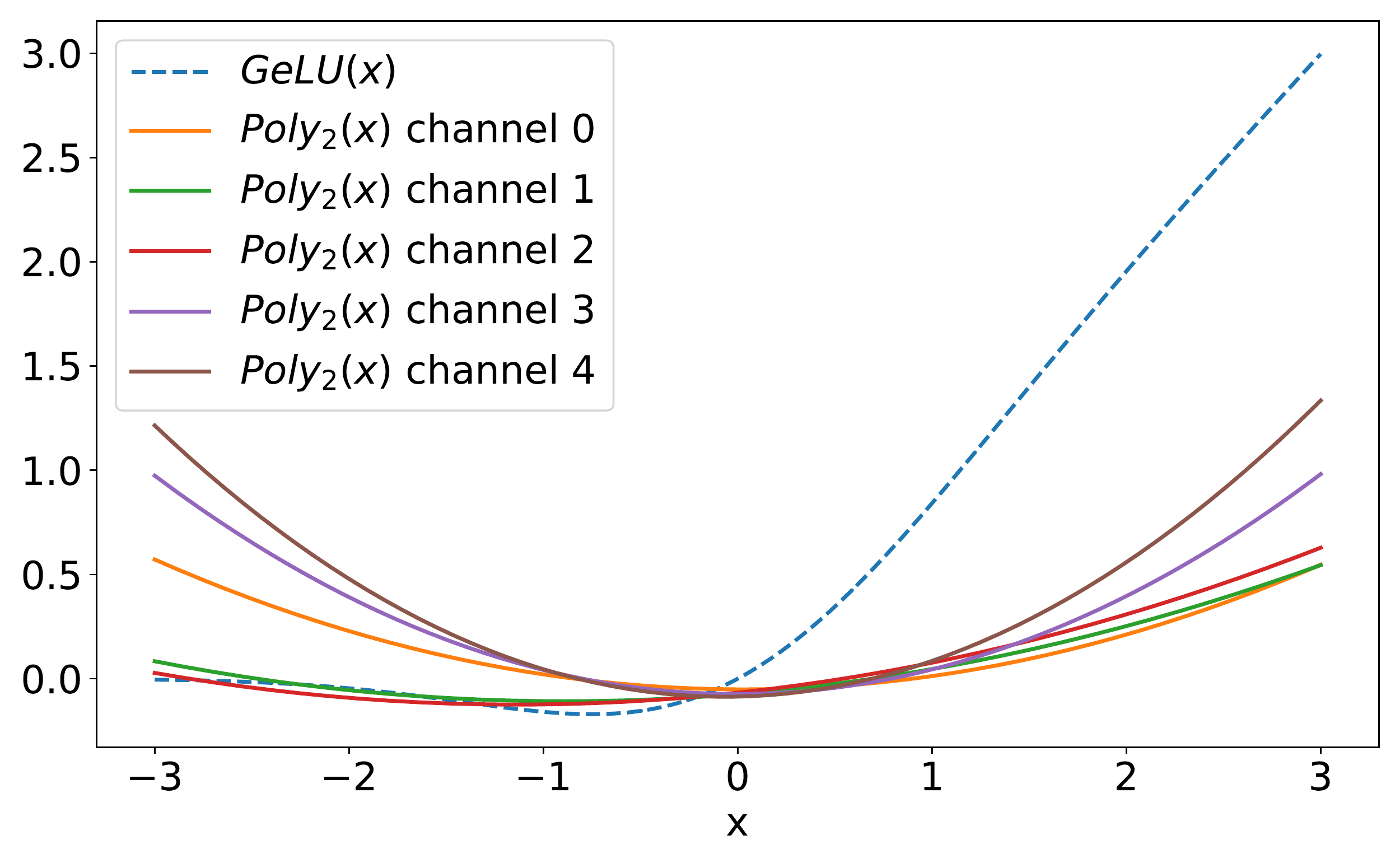}
    \end{subfigure}

\caption{
 \textbf{ The $\mathrm{Poly}_{2}(x)$ activation function.} In the left panel, we see how the coefficients are initialized to fit GeLU in range $\left[ \sqrt{2},\ -\sqrt{2} \right]$ (dashed lines).
  In the right panel we see how, in the trained model, the activation function may change significantly and converge to a different function in each channel.
  }
\label{fig:plot_poly_channels}
\end{figure}

The Polynomial activation function is a point-wise polynomial:
\begin{equation}
    \mathrm{Poly}_2(x)=a_0+a_1x+a_2x^2 \,,
\end{equation}
where the coefficients $\{a_0,a_1,a_2\}$ are trainable parameters, which are shared per-channel.
They were initialized by fitting this function to the GeLU, as proposed by \cite{Gottemukkula2019POLYNOMIALFUNCTIONS}, to function as an approximation to the original activation function ConvNeXt works well with. This initialization gives the function presented in \Cref{fig:plot_poly_channels}.
Yet, the activation function may converge to a completely different function by the end of the training. 
Moreover, it may differ significantly between different layers and between different channels in the same layer. 
\Cref{fig:plot_poly_channels} shows the final activation function for five different channels in the first block of a trained model.

\begin{figure}[ht]
\centering
\includegraphics[width=0.45\linewidth]{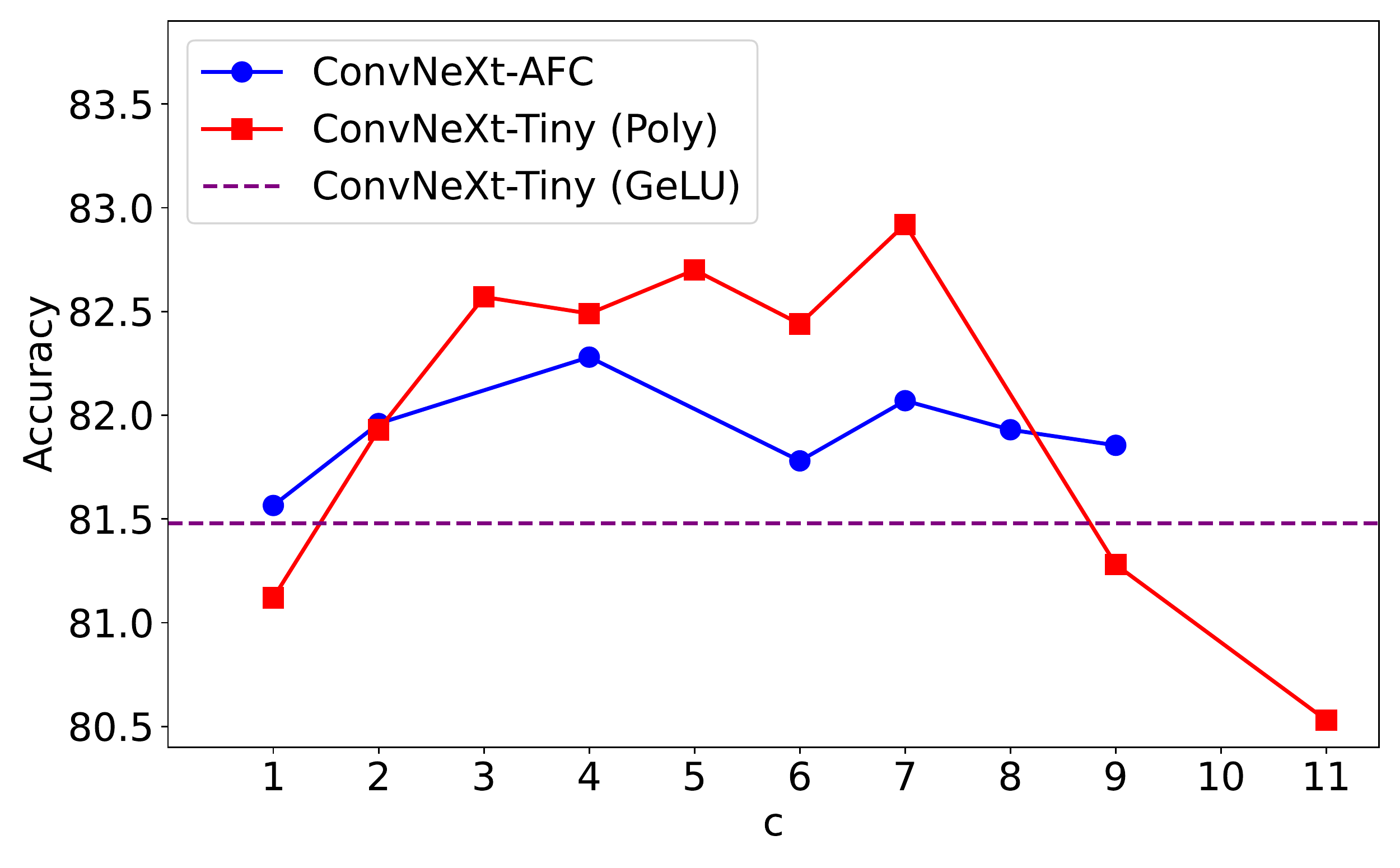}
\caption{
  \textbf{Test accuracy on ImageNet200, non-cyclic convolutions, GeLU and polynomial ConvNeXt with different scales.} Scaling with $c = 7$ gave the best results in the scope of our search.}
\label{fig:polynomial-scale-results-imnet200}
\end{figure}

\subsection{Activation scaling}
During our experiments, we found out that scaling the inputs and outputs of the activation function, regardless of the coefficients themselves, may change the final model results.
Thus, we used the activation function:
\begin{equation}
    \mathrm{Poly}_{c} \left( x \right) =  c \mathrm{Poly}_2(c x)
\end{equation}
and were looking for the optimal scale $c$. 
This scaling factor can effectively be seen as a scaling factor of the weights initialization of the pointwise convolution layers before and after the activation function. 
In \cref{fig:polynomial-scale-results-imnet200} we ran a scan over different scale factors and compared the results of the non-cyclic convolution polynomial model, without additional alias-free modifications (ConvNeXt-Tiny), and on the final model (ConvNeXt-AFC). 
The scan was run on ``ImageNet200'', a subset of ImageNet consisting of 200 classes.
We compare the results to the results of non-cyclic convolution GeLU model. 
In the scope of our search, the best result was achieved with scale $c = 7$ for ConvNeXt-Tiny and scale $c=4$ for ConvNeXt-AFC. 
We used scale $c=7$ for the rest of the polynomial models, as it achieved slightly better results on the full dataset.
\section{Formal shift-invariance proofs}
\subsection{Alias-Free polynomial activation function} \label{sec:shift-invariance-proof}

In the paper, we presented a new alias-free activation function (described in \Cref{algo:af-poly}) that, together with alias-free downsampling layers, can completely solve the aliasing problem, and lead to perfectly shift-invariant CNNs. The validity of this solution relies on the following facts:
\begin{enumerate}
    \item \cref{prop:af-poly-invariance} in the paper (which we formally prove below) ensures the activations are shift-equivariant w.r.t.~continuous domain.
    \item Convolution and alias-free downsampling layers are indeed shift-equivariant w.r.t.~continuous domain (e.g., see proof in \citep{Karras2021Alias-FreeNetworks}).
    \item A composition of functions which are shift-equivariant w.r.t.~continuous domain remains shift-equivariant w.r.t.~continuous domain. 
    \item Shift-invariance w.r.t.~continuous domain is implied from shift-equivariance w.r.t.~continuous domain, as was shown in \Cref{prop:cnn_equivariance_invariance}  of the paper.  
\end{enumerate}

\begin{figure*}[t]
  \centering
      \includegraphics[width=0.9\linewidth]{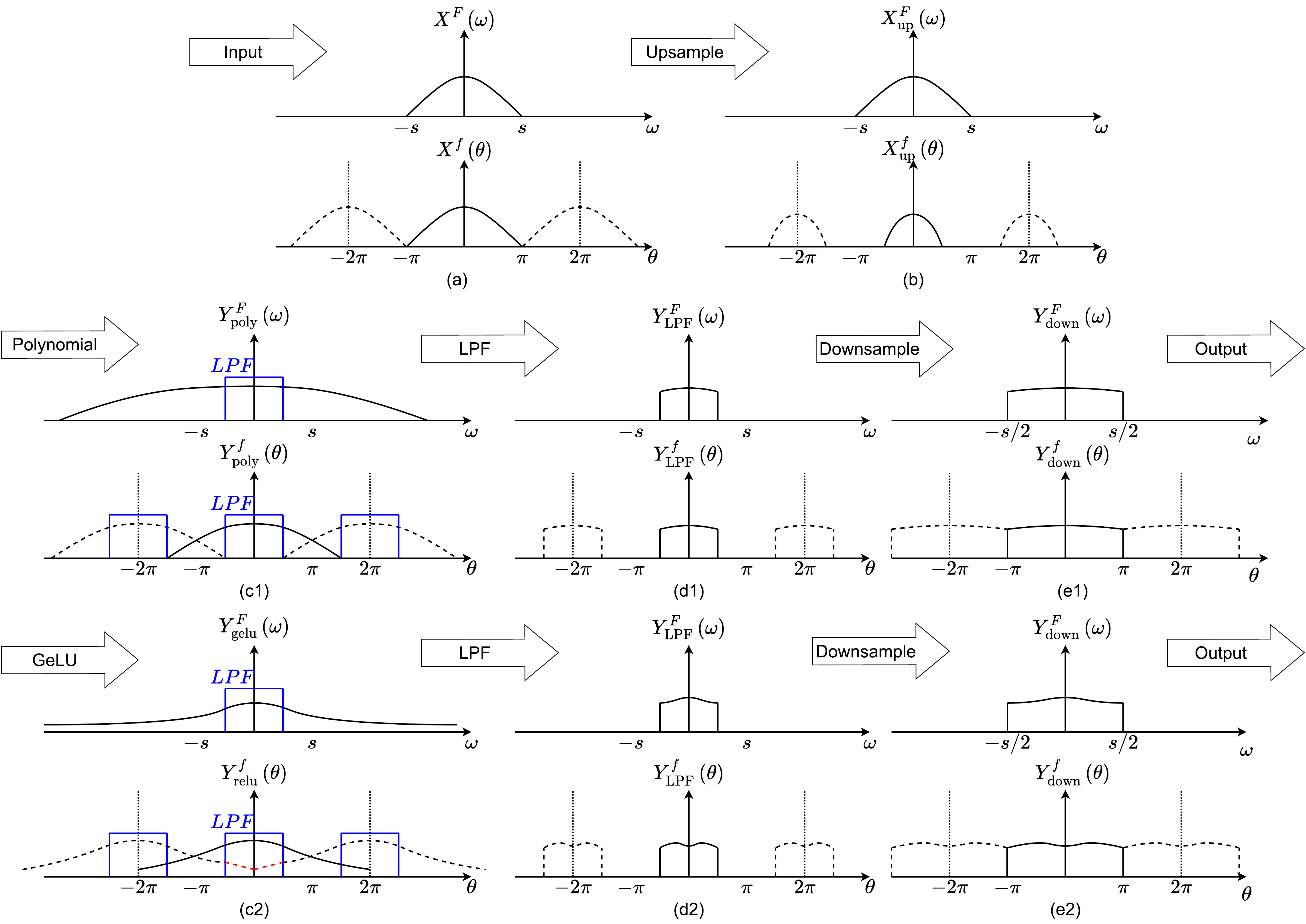}

   \caption{\textbf{A demonstration of the proposed non-linearities in the frequency domain.} The top plot at each panel represents the signal in the continuous domain, and the bottom represents the discrete domain.
   Where the input (a) is upsampled it shrinks its frequency response, expanding the allowed frequencies (b).
   Applying the polynomial activation expands the frequency response support by as factor $d$, without causing aliasing in the relevant frequencies (c.1). Thus, the discrete signal remains a faithful representation of the continuous signal after applying LPF (d1) and downsample back to the same spatial size (d2). 
   However, applying GeLU expands the support infinitely (c.2). This leads to an aliasing effect --- interference in the relevant frequencies marked in red in (c2). This causes the discrete signal not to be a correct representation of the continuous one, after LPF (d2) and downsampling (e2).}
   
   \label{fig:poly-alias-append}
\end{figure*}

Therefore, all that remains is to prove \cref{prop:af-poly-invariance}. For convenience (to help visualize the proof), we show \Cref{fig:poly-alias} of the paper again (see \cref{fig:poly-alias-append}).

\paragraph{Proof (\cref{prop:af-poly-invariance})}

We assume that the input $x$ is sampled from $x_{c}$, a $\frac{1}{T}$-band-limited signal at sample rate T, i.e. 
\begin{equation}
x\left[n\right]=x_{c}\left(nT\right)    .
\end{equation}
We denote the DTFT of $x\left[n\right]$ as:
\begin{equation}
    X^{f} \left( \theta \right) = \sum_{n = -\infty}^{\infty} x \left[ n \right] e^{-j\theta n} ,
\end{equation}
and the CTFT of $x_{c}$ as:
\begin{equation}\label{eq:ctft}
    X^{F}\left(\omega\right) = \int_{-\infty}^{\infty} x \left( t \right) e^{-j2\pi \omega t} \,dt .
\end{equation}
In addition, we define a reconstruction operator as a sinc interpolation of the discrete signal:
\begin{equation}
    \mathrm{Recon} \left( x \left[n \right] \right) \left( t \right) = \sum_{n\in \mathbb{Z}} x \left[ n \right] \mathrm{sinc} \left( \frac{t - nT}{T}\right) \,.
\end{equation}
It implies that:
\begin{equation}
    x_c \left(t \right) = \mathrm{Recon} \left( x \left[ n \right] \right) (t)\,.
\end{equation}
For easing the proof notation, we denote applying \Cref{algo:af-poly} as a whole on $x \left[ n \right]$ as $f\left( x \left[ n \right] \right)$.

A well-known relation between $X^{f} \left( \theta \right)$ and $X^{F} \left( \omega \right)$ for any continuous signal and its discrete representation is:
\begin{equation}
X^{f}\left(\theta\right)=\frac{1}{T}\sum_{k=-\infty}^{\infty}X^{F}\left(\frac{\theta+2\pi k}{T}\right) \,.
\end{equation}
This relation is represented in \Cref{fig:poly-alias-append} (a).
Since the support  of $X^{F}$ is limited by $\frac{1}{T}$, we can express
$X^{f}\left(\theta\right)$ in the frequency range $\theta\in\left[-\pi,\ \pi\right]$
as :
\begin{equation}
X^{f}\left(\theta\right)=X^{F}\left(\frac{\theta}{T}\right)    \,.
\end{equation}
From now on we will look at the DTFT domain in the range $\theta\in\left[-\pi,\ \pi\right]$,
since the effects on the rest of the replications are equal, i.e.:
\begin{equation}
\forall k\in\mathbb{Z},\ \forall\theta\in\left[-\pi,\ \pi\right]:\ X^{f}\left(\theta+2\pi k\right)=X^{f}\left(\theta\right)    \,;
\end{equation} 
this expression is true also for the DTFT of every signal from now on.

We will show that the operation presented in \Cref{algo:af-poly} on $x\left[n\right]$
is equivalent to applying polynomial activation in the continuous domain, following an LPF, i.e.
\begin{align}
f \left( x \left[ n \right] \right) \left[ n \right] &=
\mathrm{LPF}_{\frac{2}{d+1}}\left(\mathrm{Poly}_{d}\left(  x_{c} \left( t \right) \right)\right) \left( nT \right)  \\ &=
\mathrm{LPF}_{\frac{2}{d+1}}\left(\mathrm{Poly}_{d}\left( \mathrm{Recon} \left( x \left[ n \right] \right) \right)\right) \left( nT \right)   \,.
\end{align}

At step 1 of the algorithm, the signal is upsampled using sinc interpolation ($x_{\mathrm{up}}\gets\text{\ensuremath{\mathrm{Upsample}}}_{\frac{d+1}{2}}\left(x\right)$),
giving the expression 
\begin{equation}
x_{\mathrm{up}}\left[m\right]=\sum_{n\in\mathbb{Z}}x\left[n\right] \mathrm{sinc}\left( \frac{m - nI}{I} \right) \,,
\end{equation}
where $I=\frac{d+1}{2}$.

The frequency response of upsampling is a contraction in the frequency axis:
\begin{equation}
X_{\mathrm{up}}^{f}\left(\theta\right)=X^{f}\left(\theta I\right)=\begin{cases}
X^{f}\left(\theta I\right) & \left|\theta\right|\leq\pi/I\\
0 & \left|\theta\right|>\pi/I 
\end{cases}    \,,
\end{equation}
which indeed represents a sample of the continuous signal at the rate $IT$, as can be seen in \cref{fig:poly-alias-append}(b).

At step 2 of the algorithm, $\mathrm{Poly}_{d}$ is applied on $x_{\mathrm{up}}$, giving $y_{\mathrm{poly}}$.
From the duality of multiplication and convolution in spatial and Fourier domains, we get that:
\begin{equation}
Y_{\mathrm{poly}}^{f} \left( \theta \right) = a_{0}+\sum_{k=1}^{d} a_{k} \left( \frac{1}{2 \pi} \right)^{k} \underbrace{X_{\mathrm{up}}^{f}\ast...\ast X_{\mathrm{up}}^{f}}_{k\ \text{times}} \left( \theta \right) \,.  
\end{equation}
Without loss of generality, we can assume for simplicity that $\mathrm{Poly}_{d}\left(x\right)=x^{d}$ and omit the constant factor $\left( \frac{1}{2 \pi} \right)^{k}$,
since the frequency expansion is determined solely by the highest degree of the polynomial.
Since the support of $\mathrm{Poly}_{d}\left(x\right)=x^{d}$
equals to the support of $x$ multiplied by $d$, and since the input support was contracted at factor $I=\frac{d+1}{2}$, we get that the support of $\underbrace{X_{\mathrm{up}}^{f}\ast...\ast X_{\mathrm{up}}^{f}}_{d\ \text{times}}$
is $\frac{\pi d}{I}$.

We get that the polynomial output support is:
\begin{equation} \label{eq:up-poly-support}
\frac{\pi d}{I}=\frac{2\pi d}{d+1}>\pi    \,,
\end{equation}
therefore aliasing occurs. The extension of the support beyond the range $\theta \in \left[ -\pi,\ \pi \right]$ is :
\begin{equation}
\frac{2\pi d}{d+1}-\pi=\frac{\pi d-\pi}{d+1}=\pi-\frac{2\pi}{d+1}=\pi-\frac{\pi}{I}    
\end{equation}
Hence, the replications due to the aliasing do not affect the frequency domain
of $\left|\theta\right|\leq\pi/I$, i.e.:
\begin{equation} \label{eq:up-poly-dtft}
Y_{\mathrm{poly}}^{f} \left( \theta \right) =\begin{cases}
\underbrace{X_{\mathrm{up}}^{f}\ast...\ast X_{\mathrm{up}}^{f}}_{d\ \text{times}}\left(\theta I\right) & \left|\theta\right|\leq\pi/I\\
\underbrace{X_{\mathrm{up}}^{f}\ast...\ast X_{\mathrm{up}}^{f}}_{d\ \text{times}}\left(\theta I\right)+\underbrace{X_{\mathrm{up}}^{f}\ast...\ast X_{\mathrm{up}}^{f}}_{d\ \text{times}}\left(\theta I-2\pi\right) & \theta>\pi/I\\
\underbrace{X_{\mathrm{up}}^{f}\ast...\ast X_{\mathrm{up}}^{f}}_{d\ \text{times}}\left(\theta I\right)+\underbrace{X_{\mathrm{up}}^{f}\ast...\ast X_{\mathrm{up}}^{f}}_{d\ \text{times}}\left(\theta I+2\pi\right) & \theta<-\pi/I
\end{cases}
\,,
\end{equation}
where the summations in the two bottom cases represent aliasing caused by the expansion of the near replications.
This partial aliasing effect caused by the polynomial is presented in \cref{fig:poly-alias-append}(c1).

At step 3, we use an $\mathrm{LPF}_{1/I}$, thus we eliminate all the
aliased frequencies, and get:
\begin{equation}
Y_{\mathrm{LPF}}^{f} \left( \theta \right) = \begin{cases}
\underbrace{X_{\mathrm{up}}^{f}\ast...\ast X_{\mathrm{up}}^{f}}_{d\ \text{times}}\left(\theta I\right) & \left|\theta\right|\leq\pi/I\\
0 & \left|\theta\right|>\pi/I
\end{cases} \,,
\end{equation}
which can be seen in \cref{fig:poly-alias-append}(d1).

At step 4, applying $\mathrm{Downsample}_{I}$ expands the frequency
domain, so we get 
\begin{equation}
Y^{f} \left( \theta \right) =\underbrace{X^{f}\ast...\ast X^{f}}_{d\ \text{times}}\left(\theta\right),\ \theta\in\left[-\pi,\pi\right] \,.
\end{equation}
We note again that this expression is true for the domain $\theta\in\left[-\pi,\pi\right]$, specifically because the actual support of $\underbrace{X^{f}\ast...\ast X^{f}}_{d\ \text{times}}\left(\theta\right)$
is larger. However, the frequencies beyond this range were eliminated
by the LPF, as can be seen in \cref{fig:poly-alias-append}(e1).

Recalling again that $X^{f}\left(\theta\right)=X^{F}\left(\frac{\theta}{T}\right)$,
we get that the CTFT of the continuous signal of the final expression $y$ is 
\begin{equation}
Y^{F}\left(\omega\right)=\begin{cases}
\underbrace{X^{F}\ast...\ast X^{F}}_{d\ \text{times}}\left(\omega\right) & \left|\omega\right|\leq\frac{1}{T}\\
0 & \left|\omega\right|>\frac{1}{T} \,,
\end{cases}
\end{equation}
which is equivalent to the signal we would get by applying $\mathrm{LPF}_{1/I}\left(\mathrm{Poly}_{d}\left(\cdot\right)\right)$
on $x_{c}$.

Shift-equivariance w.r.t.~continuous domain stems from this equivalence because we get that
\begin{equation}
f \left( x \left[ n \right] \right) \left[ n \right] =  \mathrm{LPF}_{\frac{2}{d+1}}\left(\mathrm{Poly}_{d}\left( \mathrm{Recon} \left( x \left[ n \right] \right) \right)\right) \left( nT \right)
\end{equation}

\begin{align}\label{eq:translation-continuous-representation}
    \Rightarrow  f \left( \tau x \left[ n \right] \right) \left[ n \right] &=  
    \mathrm{LPF}_{\frac{2}{d+1}}\left(\mathrm{Poly}_{d}\left( \mathrm{Recon} \left( \tau x \left[ n \right] \right) \right)\right) \left( nT \right)   \\ 
    &=    \mathrm{LPF}_{\frac{2}{d+1}}\left(\mathrm{Poly}_{d}\left( \mathrm{Recon} \left( x \left[ n \right] \right) \right)\right) \left( nT + \Delta \right) \label{eq:translation-continuous-representation-1}\\
     &= \tau f \left( x \left[ n \right] \right) \left[ n \right] \,.
\end{align}
    

The transition in \cref{eq:translation-continuous-representation-1} is justified due to shift-equivariance w.r.t.~continuous domain of reconstruction and alias-free downsample operators, and shift-equivariance of point-wise operations in the continuous domain.

\subsection{LPF-Poly} \label{sec:lpf-poly-append}
In \Cref{sec:shift-invariance-proof} we showed that our polynomial activation function, which is derived in \Cref{algo:af-poly}, is alias-free for any polynomial. 
Specifically, as can be seen in \Cref{algo:af-poly}, the required upsample rate to avoid aliasing is dependent on the polynomial degree and equals to $\frac{d+1}{2}$, where $d$ is the polynomial degree.
In this section, we generalize this concept to cases where we would like to avoid upsampling, e.g. in layers where the channels have large spatial extents, where it is too computationally expensive. 
In addition, we show that $\mathrm{LPFPoly_2}$ which was presented in the paper is indeed alias-free and shift-equivariant w.r.t.~continuous domain, using the proof concept regarding \Cref{algo:af-poly}.

We defined LPF-Poly as:
\begin{equation}
    \mathrm{LPFPoly}_2 \left( x \left[ n \right] \right) \left[n \right] = a_0 + a_1 x\left[n \right] + x\left[n \right] \cdot \mathrm{LPF}_{c}\left(x\left[n \right] \right)\left[n \right] \,.
\end{equation}
Note that $c$ is a real number in the range $(0,1)$, representing the LPF's cutoff ratio. 
In addition, note that in the paper we omitted some of the $ \left[ n \right]$ in the notation for more compact writing. 
The output support is implied by the component of $x \cdot \mathrm{LPF}_{c}\left(x \right)$, and, similarly to the computation in \cref{eq:up-poly-support}, is equal to $\left(1 + c \right)\pi$.
We get that 
\begin{equation}
Y_{\mathrm{poly}}^{f} \left( \theta \right) =\begin{cases}
X^{f}\ast X_{\mathrm{\mathrm{LPF}}}^{f}\left(\theta\right) & \left|\theta\right|\leq\pi-\pi c\\
X^{f}\ast X_{\mathrm{\mathrm{LPF}}}^{f}\left(\theta\right)+X^{f}\ast X_{\mathrm{\mathrm{LPF}}}^{f}\left(\theta-2\pi\right) & \theta>\pi-\pi c\\
X^{f}\ast X_{\mathrm{\mathrm{LPF}}}^{f}\left(\theta\right)+X^{f}\ast X_{\mathrm{\mathrm{LPF}}}^{f}\left(\theta+2\pi\right) & \theta<-\left(\pi- \pi c\right)
\end{cases}
\,,    
\end{equation}
which means that the range $\left|\theta\right|\leq\pi(1-c)$ is alias-free. This was achieved without the need of upsampling.
Next, applying $\mathrm{LPF}_{1-c}$ gives:
\begin{equation}
Y_{\mathrm{LPF}}^{f} \left( \theta \right)=\begin{cases}
X^{f}\ast X_{\mathrm{\mathrm{LPF}}}^{f}\left(\theta\right) & \left|\theta\right|\leq\pi-\pi c\\
0                                                           & \left|\theta\right| > \pi-\pi c\\
\end{cases} \,,
\end{equation}
hence, the final output is alias-free. Shift-equivariance w.r.t.~continuous domain is derived from this, similarly to \cref{eq:translation-continuous-representation}.

\section{Implementation} \label{sec:implementation-appendix}

Our theoretical results regarding discrete representation of continuous signals are based on infinite signals, which may seem impractical to real models which work on finite images. However, the results apply in a setting in which we assume that the continuous signals are periodic, and we finitely sample a single period. These assumptions practically limit our discussion to robustness to circular translations, which is the same setting that was considered in previous works \citep{Zhang2019MakingAgain, Chaman2020TrulyNetworks}.
Next, we explain our implementation for the ``ideal LPF'', which was used in BlurPool and phase 3 of \Cref{algo:af-poly}, and for the ``reconstruction filter'', which was used in $\mathrm{Upsample}$ in step 1 of \Cref{algo:af-poly}.
Both of these filters can be implemented using multiplication in the Fourier domain, working with DFT, which is defined for a finite signal with length $N$ as
\begin{equation} \label{eq:dft}
    X^{D}[k] = \mathrm{DFT} \left( x \left[ n \right] \right) \left[ k \right] = \sum_{n=0}^{N-1} x \left[ n \right] e^{-\frac{j 2 \pi}{N}kn } \,.
\end{equation}
Similarly, the inverse of DFT is defined as:
\begin{equation} \label{eq:dft-inv}
    x \left[ n \right]=\mathrm{IDFT} \left( X^D \left[ k \right] \right) \left[ n \right] = \frac{1}{N} \sum_{k=0}^{N-1} X^{D} \left[ k \right] e^{\frac{j 2 \pi}{N}kn } \,.
\end{equation}
For simplicity, all our derivations are for 1-D signals. Our derivations trivially apply to the 2-D case by applying the filters separately on rows and on columns.
\paragraph{LPF}
\label{par:lpf}
We used a low-pass filter wherever it was necessary to prevent aliasing due to downsampling, namely in BlurPool layers that replace strided convolutions, and alias-free polynomial activations, before subsampling the polynomial results (\cref{algo:af-poly}). We used an ``ideal filter'', i.e.~a filter that eliminates all the frequencies above the cutoff ratio. Practically, this kind of filter can be implemented using multiplication in DFT domain: 
\begin{equation}
    \mathrm{LPF}_{\mathrm{cutoff},c}  \left( x \left[ n \right] \right) \left[ n \right] =   \mathrm{IDFT} \left(\mathrm{DFT} \left(x \left[ n \right] \right) \left[ k \right] H^{D}_{\mathrm{cutoff},c}\left[ k \right] \right) \left[ n \right]\,,
\end{equation}
where $H^{D}_{\mathrm{cutoff, c}}$ is a ``rectangle filter'' defined for spatial dimension $N$  and cutoff ratio $c \in \left[ 0,\ 1\right]$ as 

\begin{equation}
H^{D}_{\mathrm{cutoff,c}} \left[ k \right]=\begin{cases}
1 & 0\leq k<\frac{N}{2}c\,,\\
0 & \frac{N}{2}c \leq k \leq N - \frac{N}{2}c\,,\\
1 & N - \frac{N}{2}c < k\leq N-1\,.
\end{cases} \,.
\end{equation}

\paragraph{Downsampling}
As mentioned above, all downsampling operations were performed in an alias-free manner, using low-pass filters before subsampling. 
For subsampling at factor $s$, we used LPF with cutoff ratio $c = \frac{1}{s}$.
Then, we used subsampling with a fixed grid:
\begin{equation}
    x_{\mathrm{down}} \left[n \right] = \mathrm{LPF}_{1/s} \left( x \left[ n \right] \right) \left[ sn \right]
\end{equation}

\paragraph{Upsampling}

In the proof of Algorithm 1, we assume we use ``ideal upsample'', which can be interpreted as a re-sampling in a higher rate of the continuous signal, which was restored using sinc interpolation:
\begin{equation}
    x_{\mathrm{up}_{I}} \left[ m \right] = \sum_{n} x \left[n \right] \mathrm{sinc} \left( \frac{m - nI}{I} \right)
\end{equation}
In practice, and specifically in a finite signal case, upsampling is performed in two steps:
First, we use zero padding and get the intermediate signal 
\begin{equation}
    x_{z}\left[m \right] = \begin{cases}
        x \left[ \frac{m}{I} \right ] &     m = kI\, , \\
       0                               & \text{otherwise.}
    \end{cases}
\end{equation}
Then, the zero-padded signal is convolved with sinc interpolation kernel. This step is equivalent to multiplication in Fourier domain with a rectangle, similarly to the LPF implementation.
\begin{equation}
\mathrm{Upsample}_{I}  \left( x \left[ n \right] \right)\left[ n \right] =   \mathrm{IDFT} \left( \mathrm{DFT} \left( x_z \left[ n \right] \right)\left[ k \right] H^{D}_{\mathrm{upsample}, I}\left[ k \right] \right) \left[ n \right]\,,
\end{equation}
Practically we used the following upsample kernel for a signal with spatial dimension $N$.
For even $N$:
\begin{equation}
    H^{D}_{\mathrm{upsample},I} \left[ k \right]=\begin{cases}
1 & 0\leq k<\frac{N}{2},\\
1 & N\left(I - \frac{1}{2}\right)+1\leq k\leq IN-1\\
0.5 & k=\frac{N}{2},k=N\left(I - \frac{1}{2}\right)\\
0 & \text{else}
\end{cases}
\end{equation}
For odd N:
\begin{equation}
H^{D}_{\mathrm{upsample}, I}\left[k\right]=\begin{cases}
1 & 0\leq k<\lfloor\frac{N}{2}\rfloor\\
1 & \lceil N\left(I - \frac{1}{2}\right) \rceil\leq k\leq2N-1\\
0 & \text{else}
\end{cases}	
\end{equation}
The reason for that is that in practice, we cannot assume the Nyquist condition holds. Specifically, for a finite signal $x\left[ n \right]$ with an even size $N$, we cannot assume that $X^{D} \left[ \frac{N}{2}\right] = X^{D} \left[ \frac{3N}{2}\right] = 0$. Note that for signals with even length, the $\frac{N}{2}$ component in the DFT domain represents the continual frequency of $\frac{\pi}{T}$, and thus, due to aliasing effect, we have
 $X^{D} \left[ \frac{N}{2}\right] = X^{D} \left[ \frac{3N}{2}\right] = X^F \left( \frac{\pi}{T} \right) + X^F \left( \frac{-\pi}{T} \right)$. 
 For a representation of the continuous signal with a higher sampling rate (e.g. the upsampled signal), the overlap in this frequency would not happen, hence we multiply this component by $\frac{1}{2}$ to get $\frac{1}{2} \left( X^F \left( \frac{\pi}{T} \right) + X^F \left( \frac{-\pi}{T} \right) \right) = X^F \left( \frac{\pi}{T} \right)$.
In this equation we use the assumption that $X^F \left( \frac{\pi}{T} \right) \in \mathbb{R}$. 
For a real signal The CTFT is conjugate symmetric, meaning 
$X^F \left( \frac{\pi}{T} \right) = X^F \left( \frac{-\pi}{T} \right)^{*} $. 
Therefore in case $X^F \left( \frac{\pi}{T} \right)$ has an imaginary component, it cannot be retrieved from the sum.
 A more detailed proof of this is given in appendix \cref{sec:implementation-proofs} below.

\section{Implementation proofs}
\label{sec:implementation-proofs}
 In the following section, we provide formal proofs for the correctness of the filters presented in \cref{sec:implementation-appendix} In the setting of finite discrete signals, where we assume the continuous domain signals are periodic.
 For simplicity to the reader, we prove the correctness for 1-dimensional signal, and for upsampling at factor 2. The proofs can be easily generalized to 2D signals and a higher upsampling rate.

\subsection{Definitions}\label{sec:implementation-proofs-defs}
Let $x\left(t\right)$ be a band-limited continuous signal, with CTFT  $X^{F}\left(\omega\right)$ (as defined in \cref{eq:ctft}). $x\left(t\right)$ is periodic with period $NT$, i.e. $x\left(NT+t\right)=x\left(t\right)$. We define discrete sampling as:
\begin{equation}
    x\left[n\right]=x\left(nT\right)\,,    
\end{equation}
and define a finite sampling as taking only one period of the discrete signal, i.e. 
\begin{equation}
    x_{N}\left[n\right]=\left\{ x_{0},...,x_{N-1}\right\} =\left\{ x\left(0\right),x\left(T\right),...,x\left(\left(N-1\right)T\right)\right\} \,.    
\end{equation}

\subsection{Upsample}
As noted in \cref{sec:implementation-appendix}, we assume that $X^{F}(\omega)=0\ \ \forall|\omega|>\frac{\pi}{T}$  and that $X^{F}\left( \frac{\pi}{T}\right) \in \mathbb{R}$ (This a relaxation of Nyquist condition for which $X^{F} \left( \frac{\pi}{T} \right) = 0$). We prove the validity of the following method to upsample $x_{N} \left[ n \right]$  to retrieve:
\begin{equation}\label{eq:x-2n}
    x_{2N}\left[n\right]=x\left(n\frac{T}{2}\right)=\left\{ x\left(0\right),x\left(\frac{T}{2}\right),...,x\left(\left(2N-1\right)\frac{T}{2}\right)\right\} \,.
\end{equation}
The upsample method presented in \Cref{sec:implementation-appendix} is formally shown in \Cref{algo:Up-sample}:


\begin{algorithm}[h]
\caption{Upsample}
\label{algo:Up-sample}
\begin{algorithmic}
\item {\bfseries Input:} $x_{N}\left[n\right]\in\mathbb{R}^{N}$ 
    \item $x_{z} \gets \left\{ x_{N} \left[ 0 \right],\ 0,\ x_{d} \left[ 1 \right],\ 0\ ,...,\ x_{d} \left[ N-1 \right],\ 0 \right\}$ 
    \item $x_{2N} \gets \mathrm{IDFT}\left\{ \mathrm{DFT} \left\{ x_{z}\right\} H_{2}^{D}\right\} $
    \item {\bfseries Output:} $x_{2N}\left[ n \right]$ 
    \end{algorithmic}
\end{algorithm}



\begin{claim} \label{prop:upsample-algo}
Let $x\left(t\right)$ and $x_{N}\left[n\right]$ be a continuous signal and its finite discrete representation as defined in \cref{sec:implementation-proofs-defs}.
Then the output of \Cref{algo:Up-sample} is 
\[
    x_{2N}\left[n\right]=x\left(n\frac{T}{2}\right)=\left\{ x\left(0\right),x\left(\frac{T}{2}\right),...,x\left(\left(2N-1\right)\frac{T}{2}\right)\right\} \,,
\]
using the described reconstruction filter $H^{D}_{2}$ below:

For even N:
\begin{equation}
    H^{D}_{2} \left[k \right]=\begin{cases}
        1 & 0\leq k<\frac{N}{2},\\
        1 & \frac{3N}{2}+1\leq k\leq2N-1\\
        0.5 & k=\frac{N}{2},k=\frac{3N}{2}\\
        0 & else
    \end{cases}
    \end{equation}
For odd N:
\begin{equation}
    H^{D}_{2} \left[k \right]=\begin{cases}
        1 & 0\leq k<\lfloor\frac{N}{2}\rfloor\\
        1 & \lceil\frac{3N}{2}\rceil\leq k\leq2N-1\\
        0 & else
    \end{cases}
\end{equation}
\end{claim}
\paragraph{Proof (\cref{prop:upsample-algo})}
In order to proof \cref{prop:upsample-algo}, we will show that the DFT of its output equals to $X^{D}_{2N}$, i.e.~the DFT of the signal $x_{2N}$ that is defined in \cref{eq:x-2n}.

\subsubsection{DFT of zero-padded signal}
Recall that in the first step of the \cref{algo:Up-sample} we apply zero padding on the input $x_{N}$.
For a finite discrete signal with length $N$, DFT is defined as \cref{eq:dft}:
\[
X^{D}[k]=\sum_{n=0}^{N-1}x[n]e^{-\frac{j2\pi}{N}nk}\underbrace{=}_{W_{N}\triangleq e^{\frac{j2\pi}{N}}}\sum_{n=0}^{N-1}x[n]W_{N}^{-nk}
\]

\begin{claim} \label{prop:dft-zero-pad}
Let $X_{N}^{D}$ be the DFT of the input $x_{N}$ of \Cref{algo:Up-sample} and $X_{z}^{D}$ be the DFT of the zero-padded signal $x_{z}$ in step 1 of \Cref{algo:Up-sample}. Then:
\begin{equation}
    X_{z}^{D}[k]=\begin{cases}
        X_{N}^{D}[k] & k<N\\
        X_{N}^{D}[k-N] & k\geq N
    \end{cases}\label{eq:1}
\end{equation}
\end{claim}
\paragraph{Proof (\cref{prop:dft-zero-pad})}
\begin{align}
X_{z}^{D} \left[ k \right] &= \sum_{n=0}^{2N-1} x_{z} \left[ n \right] W_{2N}^{-nk} \\
&\underset{(\ast)}{=} \sum_{n'=0}^{N-1}\left(x_{z} \left[ 2n' \right] W_{2N}^{-2n'k} + \underbrace{x_{z}[2n'+1]}_{=0\ \forall n'}W_{2N}^{-(2n'+1)k}\right) \\
 & =\sum_{n'=0}^{N-1}x_{z} \left[ 2n' \right] W_{2N}^{-2n'k}\\
 & \underset{(\ast \ast)}{=}\sum_{n'=0}^{N-1}x_{N} \left[ n' \right]W_{N}^{-n'k} \\
\end{align}
In $(\ast)$ we separated the summation to even and odd components, and in $(\ast \ast)$ we used the fact that 
\[
W_{2N}^{-2n'k}=e^{\frac{j2\pi}{2N}(-2n'k)}=e^{\frac{j2\pi}{N}(-n'k)}=W_{N}^{-n'k} \,.
\]
Note that we got a sum of $N$ components, yet $X_{z}^{D}[k]$ is defined for $k=0,1,...,2N-1$.
For $k=0,1,...,N-1$ we got the definition of DFT, meaning 
\begin{equation}
X_{z}^{D} \left[ k \right] = X_{N}^{D} \left[ k \right] \,.    
\end{equation}
For $k=N,...,2N-1$ using the property 
\[
W_{N}^{-n'(N+k)}=\underbrace{W_{N}^{-n'N}}_{=e^{\frac{j2\pi}{N}(-n'N)}=e^{-j2\pi n'}=1}W_{N}^{-n'k}=W_{N}^{-n'k}
\]
we get:
\begin{align}
X_{z}^{D} \left[ k \right] &= \sum_{n'=0}^{N-1}x_{N} \left[ n' \right] W_{N}^{-n'k} \\
&= \sum_{n'=0}^{N-1}x_{N} \left[ n' \right] W_{N}^{-n' \left( N+ \left( k-N \right) \right)} \\
&= \sum_{n'=0}^{N-1}x_{N} \left[ n' \right] W_{N}^{-n' \left( k-N \right) } \\
&= X_{N}^{D}\left[ k-N \right] \,.
\end{align}

\subsubsection{Expressing DFT with CTFT } \label{sec:dft-ctft}
In the previous section we expressed $X_{z}^{D}$ with $X_{N}^{D}$.
In the following section we will express $X_{2N}^{D}$ using $X_{N}^{D}$.
In addition, we assume that $N$ is even, and will show the other case afterwards.
First, we express the DFT of $x_{N}$ with the CTFT of $x$, by using their relations with DTFT:
\begin{align}
X_{N}^{D} \left[ k \right] &= X_{N}^{f} \left( \theta=\frac{2\pi k}{N} \right) \\
&= \frac{1}{T}\sum_{l=-\infty}^{\infty}X^{F}\left( \frac{\theta+2\pi l}{T} \right) \\
&= \frac{1}{T}\sum_{l=-\infty}^{\infty}X^{F} \left( \frac{2\pi k}{NT}+\frac{2\pi l}{T} \right) \\
& \underbrace{=}_{X^{F} \left( \omega \right) = 0\ \ \forall|\omega|>\frac{\pi}{T}}\frac{1}{T} \left( X^{F} \left( \frac{2\pi k}{NT} \right)+X^{F} \left( \frac{2\pi k}{NT}-\frac{2\pi}{T} \right) \right)
\end{align}

\begin{equation} \label{eq:x-n-dft} 
\Rightarrow X_{N}^{D}[k]=\frac{1}{T}\begin{cases}
X^{F}(\frac{2\pi k}{NT}) & 0\leq k\leq\frac{N}{2}-1\\
X^{F}(\frac{\pi}{T})+X^{F}(-\frac{\pi}{T}) & k=\frac{N}{2}\\
X^{F}(\frac{2\pi k}{NT}-\frac{2\pi}{T}) & \frac{N}{2}+1\leq k\leq N-1
\end{cases} \,.
\end{equation}
Note that the last two transitions hold considering the limited support
of $X^{F}:$
\[
\underbrace{X^{F}(\frac{2\pi k}{NT})}_{=0\ \ \forall k>\frac{N}{2}}+\underbrace{X^{F}(\frac{2\pi k}{NT}-\frac{2\pi}{T})}_{=0\ \ \forall k<\frac{N}{2}} \,.
\]
Next, we will derive a similar expression for $X_{2N}^{D}[k]$:
\begin{align}
X_{2N}^{D} \left[ k \right] &= X_{2N}^{f} \left( \theta=\frac{2\pi k}{2N} \right) \\
&= \frac{1}{T}\sum_{l=-\infty}^{\infty}X^{F}\left( \frac{\theta+2\pi l}{T/2} \right) \\
&= \frac{1}{T} \left( \underbrace{ X^{F} \left(\frac{2\pi k}{NT} \right)}_{=0\ \ \forall k>\frac{N}{2}}+\underbrace{X^{F}\left( \frac{2\pi k}{NT}-\frac{2\pi}{T/2} \right)}_{=0\ \ \forall k<\frac{3N}{2}} \right)\\
\end{align}
 We get: 
\begin{equation}
\Rightarrow X_{2N}^{D}[k]=\frac{1}{T}\begin{cases}
X^{F}(\frac{2\pi k}{NT}) & 0\leq k\leq\frac{N}{2}\\
0 & \frac{N}{2}+1\leq k\leq\frac{3N}{2}-1\\
X^{F}(\frac{2\pi k}{NT}-\frac{4\pi}{T}) & \frac{3N}{2}\leq k\leq2N-1
\end{cases}\label{eq:3}
\end{equation}

\subsubsection{Expressing $X_{2N}^{D}$ with $X_{N}^{D}$}
Considering the second step of \Cref{algo:Up-sample}, we need to show that applying the filter $H^{D}_{2}$ on $X_{z}^{D}$ yields $X_{2N}^{D}$, meaning
\[
X_{2N}^{D}\left[ k \right] = X_{z}^{D} \left[k \right] H^{D}_{2} \,.
\]
By plugging $X_{N}^{D} \left[ k \right]$ (\cref{eq:x-n-dft}) in $X_{z}^{D} \left[k \right]$ (\cref{eq:1})
we get: 
\begin{align}
X_{z}^{D} \left[k \right] &=\begin{cases}
    X_{N}^{D} \left[ k \right] & k<N\\
    X_{N}^{D}\left[k-N \right] & k\geq N
    \end{cases} \\
 &= \frac{1}{T}\begin{cases}
    X^{F} \left( \frac{2\pi k}{NT} \right) & 0\leq k\leq\frac{N}{2}-1\\
    X^{F} \left(\frac{\pi}{T} \right)+X^{F} \left(-\frac{\pi}{T} \right) & k=\frac{N}{2}\\
    X^{F}\left( \frac{2\pi k}{NT}-\frac{2\pi}{T} \right) & \frac{N}{2}+1\leq k\leq N-1\\
    X^{F} \left(\frac{2\pi \left(k-N \right)}{NT} \right) & N\leq k\leq\frac{3N}{2}-1\\
    X^{F} \left(\frac{\pi}{T}\right)+X^{F} \left(-\frac{\pi}{T}\right) & k=\frac{3N}{2}\\
    X^{F} \left(\frac{2\pi \left(k-N \right)}{NT}-\frac{2\pi}{T} \right) & \frac{3N}{2}+1\leq k\leq2N-1
\end{cases} \,.
\end{align}
Thus, by applying the filter 
\begin{equation}
H^{D}_{2} \left[ k \right]=\begin{cases}
1 & 0\leq k<\frac{N}{2},\\
1 & \frac{3N}{2}+1\leq k\leq2N-1\\
0.5 & k=\frac{N}{2},k=\frac{3N}{2}\\
0 & else
\end{cases}
\end{equation}

we get: 
\begin{equation}
H^{D}_{2} \left[ k \right] X_{z}^{D} \left[ k \right] = \frac{1}{T} \begin{cases}
X^{F} \left( \frac{2\pi k}{NT} \right)                                                              & 0\leq k\leq\frac{N}{2}-1\\
\frac{1}{2}\left(X^{F} \left(\frac{\pi}{T} \right)+X^{F} \left( -\frac{\pi}{T} \right)\right)      & k=\frac{N}{2}\\
0                                                                                                     & \frac{N}{2}+1\leq k\leq N-1\\
0                                                                                                   & N\leq k\leq\frac{3N}{2}-1\\
\frac{1}{2}\left(X^{F} \left( \frac{\pi}{T} \right)+X^{F} \left(-\frac{\pi}{T} \right)\right)                                      & k=\frac{3N}{2}\\
X^{F} \left(\frac{2\pi \left( k-N \right) }{NT}-\frac{2\pi}{T} \right)                              & \frac{3N}{2}+1\leq k\leq2N-1
\end{cases} \,.
\end{equation}

Note that for real $x(t)$, $X^{F}$ is conjugate symmetric, and we assumed  that $X^{F} \left(\frac{\pi}{T} \right) \in \mathbb{R}$. Therefore:
\begin{itemize}
\item for $k=\frac{N}{2},\frac{3N}{2}$: 
\begin{equation}
X^{F} \left(\frac{\pi}{T} \right) + X^{F} \left( -\frac{\pi}{T} \right) = 2X^{F} \left( \frac{\pi}{T} \right)=2X^{F} \left(-\frac{\pi}{T} \right) \,.
\end{equation}
\item for $\frac{3N}{2}+1\leq k\leq2N-1$: 
\begin{equation}
X_{z}^{D} \left[ k \right] = X^{F} \left( \frac{2\pi \left( k-N \right) }{NT}-\frac{2\pi}{T} \right)=X^{F} \left( \frac{2\pi k}{NT}-\frac{2\pi N}{NT}-\frac{2\pi}{T} \right)=X^{F} \left( \frac{2\pi k}{NT}-\frac{4\pi}{T} \right) \,.    
\end{equation}
\end{itemize}
Thus we get: 
\begin{align}
H^{D}_{2} \left[ k \right] X_{z}^{D} \left[ k \right] &= \frac{1}{T}\begin{cases}
X^{F} \left(\frac{2\pi k}{NT}\right)                 & 0\leq k\leq\frac{N}{2}-1\\
X^{F} \left(\frac{\pi}{T}\right)                    & k=\frac{N}{2}\\
0                                                     & \frac{N}{2}+1\leq k\leq N-1\\
0                                                   & N\leq k\leq\frac{3N}{2}-1\\
X^{F} \left( -\frac{\pi}{T} \right)                 & k=\frac{3N}{2}\\
X^{F} \left( \frac{2\pi k}{NT}-\frac{4\pi}{T} \right) & \frac{3N}{2}+1\leq k\leq2N-1
\end{cases} \\
&= \frac{1}{T}\begin{cases}
X^{F} \left( \frac{2\pi k}{NT} \right)              & 0\leq k\leq\frac{N}{2}\\
0                                               & \frac{N}{2}+1\leq k\leq\frac{3N}{2}-1\\
X^{F} \left( \frac{2\pi k}{NT}-\frac{4\pi}{T} \right) & \frac{3N}{2}\leq k\leq2N-1
\end{cases} \\
& \underset{(\ref{eq:3})}{=} X_{2N}^{D} \left[ k \right] \,.  
\end{align}

\subsubsection{Odd $N$}
By repeating the derivations of section \cref{sec:dft-ctft} for odd $N$ we get: 
\begin{equation}
X_{N}^{D} \left[k \right]=\frac{1}{T}\begin{cases}
X^{F}\left(\frac{2\pi k}{NT} \right) & 0\leq k\leq\lfloor\frac{N}{2}\rfloor\\
X^{F}\left(\frac{2\pi k}{NT}-\frac{2\pi}{T} \right) & \lceil\frac{N}{2}\rceil\leq k\leq N-1
\end{cases}\label{eq:2.1} \,.
\end{equation}
 
\begin{equation}
X_{2N}^{D}\left[k\right]=\frac{1}{T}\begin{cases}
X^{F}\left(\frac{2\pi k}{NT}\right) & 0\leq k\leq \lfloor\frac{N}{2} \rfloor\\
0 & \lceil\frac{N}{2}\rceil\leq k\leq\lfloor\frac{3N}{2}\rfloor\\
X^{F}\left(\frac{2\pi k}{NT}-\frac{4\pi}{T}\right) & \lceil\frac{3N}{2}\rceil\leq k\leq2N-1
\end{cases}\label{eq:3.1} \,.
\end{equation}
By plugging $X_{N}^{D}\left[k\right]$ (\ref{eq:2.1}) in $X_{0}^{D}\left[k\right]$ (\ref{eq:1})
we get: 
\begin{align}
X_{z}^{D}\left[k\right] & =\begin{cases}
X_{N}^{D}\left[k\right] & k<N\\
X_{N}^{D}\left[k-N\right] & k\geq N
\end{cases} \\
 &= \frac{1}{T}\begin{cases}
X^{F}\left(\frac{2\pi k}{NT}\right) & 0\leq k\leq\lfloor\frac{N}{2}\rfloor\\
X^{F}\left(\frac{2\pi k}{NT}-\frac{2\pi}{T}\right) & \lceil\frac{N}{2}\rceil\leq k\leq N-1\\
X^{F}\left(\frac{2\pi\left(k-N\right)}{NT}\right) & N\leq k\leq\lfloor\frac{3N}{2}\rfloor\\
X^{F}\left(\frac{2\pi\left(k-N\right)}{NT}-\frac{2\pi}{T}\right) & \lceil\frac{3N}{2}\rceil\leq k\leq2N-1
\end{cases} \,.
\end{align}
Then, by applying the filter 
\begin{equation}
H^{D}_{2}\left[k\right]=\begin{cases}
1 & 0\leq k<\lfloor\frac{N}{2}\rfloor,\\
1 & \lceil\frac{3N}{2}\rceil\leq k\leq2N-1\\
0 & else
\end{cases}
\end{equation}
 we get: 
\begin{align}
H^{D}_{2}\left[k\right]X_{z}^{D}\left[k\right] &= \frac{1}{T}\begin{cases}
X^{F}\left(\frac{2\pi k}{NT}\right) & 0\leq k\leq\lfloor\frac{N}{2}\rfloor\\
0 & \lceil\frac{N}{2}\rceil\leq k\leq \lfloor\frac{3N}{2}\rfloor \\
X^{F}\left(\frac{2\pi\left(k-N\right)}{NT}-\frac{2\pi}{T}\right) & \lceil\frac{3N}{2}\rceil\leq k\leq2N-1
\end{cases} \\
& \underset{(\ref{eq:3.1})}{=} X_{2N}^{D}\left[k\right] \,.
\end{align}

\subsection{LPF}
We use an ``ideal LPF'' before downsampling in BlurPool layers and in alias-free polynomial activations.
As mentioned in \cref{sec:shift-invariance-proof} ``ideal LPF'' in the context of subsampling of infinite discrete signals is a filter that completely eliminates all the frequencies beyond the Nyquist condition.
E.g. in case of subsampling in factor $I$, i.e. 
\[
y \left[ n \right] = x \left[ I n \right] \,,
\]
an ``ideal LPF'' in DTFT domain is implemented by multiplication with the filter:
\[
H^{f}_{1/I} \left(\theta \right) = \begin{cases}
    1  &    \left| \theta \right| < \frac{\pi}{I} \,, \\
    0  &     \left| \theta \right| \geq \frac{\pi}{I}
\end{cases}
\]
When considering the continuous domain, we expect the results to be equal to the discrete representation of the continuous signal after applying ``ideal LPF'', that is multiplication in CTFT domain with the filter 
\[
H^{F}_{1/I} \left(\theta \right) = \begin{cases}
    1  &    \left| \theta \right| < \frac{\pi}{TI}  \\
    0  &     \left| \theta \right| \geq \frac{\pi}{TI}
\end{cases} \,,
\]
where $T$ is the sample rate of $x \left[ n \right]$.
\begin{claim} \label{prop:ideal-lpf}
Let $x\left(t\right)$ and $x_{N}\left[n\right]$ be a continuous signal and its finite discrete representation as defined in \cref{sec:implementation-proofs-defs}.
In addition, let $x_{\mathrm{LPF}} \left( t \right)$ be the continuous signal received by applying $H^{F}_{1/I}$ on $x \left(t \right)$ in the continuous domain, i.e.~:
\begin{equation} \label{eq:x-lpf-ctft}
X_{\mathrm{LPF}}^{F}\left(\omega\right)	=\begin{cases}
X^{F}\left(\omega\right) & \left|\omega\right|<\frac{\pi}{TI}\\
0 & \left|\omega\right|\geq\frac{\pi}{TI}
\end{cases}
\end{equation}
In addition, define $x_{\mathrm{LPF}} \left( t \right)$ discrete representation as: 
\[
x_{\mathrm{LPF}} \left[ n \right] = \left\{ x_{\mathrm{LPF}}\left(0\right),\ x_{\mathrm{LPF}}\left(T\right),...,\ x_{\mathrm{LPF}}\left(\left(N-1\right)T\right)\right\} \,.    
\]
Then applying $\mathrm{LPF}_{1/I}$  on $x_{N}$ gives
 \[
 \mathrm{LPF}_{1/I} \left( x_{N} \right) = x_{\mathrm{LPF}}\left[ n\right]
 \]
where $\mathrm{LPF}_{1/I}$ is defined as multiplication in DFT domain with
\[
H^D_{1/I} \left[ k \right]=\begin{cases}
1 & 0\leq k<\frac{N}{2I}\,,\\
0 & \frac{N}{2I}\leq k \leq N - \frac{N}{2I}\,,\\
1 & N - \frac{N}{2I} < k\leq N-1\,.
\end{cases} \,.
\]
\end{claim}

\paragraph{proof (\cref{prop:ideal-lpf})}
Similarly to \cref{sec:dft-ctft}, using the relations between DFT, DTFT and CTFT we get:

\begin{align}
X_{\mathrm{LPF}}^{D}\left[k\right]	&=  X_{\mathrm{LPF}}^{f}\left(\theta=\frac{2\pi k}{N}\right) \\
&=  \frac{1}{T}\sum_{l=-\infty}^{\infty}X_{\mathrm{LPF}}^{F}\left(\frac{\theta+2\pi l}{T}\right) \\
&= \frac{1}{T}\sum_{l=-\infty}^{\infty}X_{\mathrm{LPF}}^{F}\left(\frac{2\pi k}{NT}+\frac{2\pi l}{T}\right) \\
& \underbrace{=}_{X^{F}(\omega)=0\ \ \forall|\omega|>\frac{\pi}{T}}\frac{1}{T}\left(X_{\mathrm{LPF}}^{F}\left(\frac{2\pi k}{NT}\right)+X_{\mathrm{LPF}}^{F}\left(\frac{2\pi k}{NT}-\frac{2\pi}{T}\right)\right)    \,.
\end{align}
By plugging \cref{eq:x-lpf-ctft} we get:
\begin{equation}
X_{\mathrm{LPF}}^{D}\left[k\right]=\frac{1}{T}\begin{cases}
X^{F}\left(\frac{2\pi k}{2T}\right) & 0\leq k<\frac{N}{2I}\\
0 & \frac{N}{2I}\leq k\leq N-\frac{N}{2I}\\
X_{\mathrm{}}^{F}\left(\frac{2\pi k}{NT}-\frac{2\pi}{T}\right) & N-\frac{N}{2I}<k\leq N-1 \,.
\end{cases}
\end{equation}
Recall that $x_{N}$ satisfies (\cref{eq:x-n-dft}): 
\[ 
X_{\mathrm{N}}^{D}\left[k\right]=\frac{1}{T}\begin{cases}
X^{F}\left(\frac{2\pi k}{2T}\right) & 0\leq k<\frac{N}{2}-1\\
X^{F}\left(\frac{\pi}{T}\right)+X^{F}\left(-\frac{\pi}{T}\right) & k=\frac{N}{2}\\
X_{\mathrm{}}^{F}\left(\frac{2\pi k}{NT}-\frac{2\pi}{T}\right) & \frac{N}{2}<k\leq N-1
\end{cases} \,;
\]
thus we get 
\begin{equation}
X_{\mathrm{LPF}}^{D}\left[k\right]=H_{1/I}^{D}\left[k\right]X^{D}\left[k\right] \,.
\end{equation}

\end{document}